\documentclass{article}

     \usepackage[preprint,nonatbib]{neurips_2020}

\usepackage[utf8]{inputenc} 
\usepackage[T1]{fontenc}    
\usepackage{hyperref}       
\usepackage{url}            
\usepackage{booktabs}       
\usepackage{amsfonts}       
\usepackage{nicefrac}       
\usepackage{microtype}      

\usepackage{graphicx}
\usepackage{subfigure}
\usepackage{algorithm}
\usepackage{algorithmic}
\usepackage[square,numbers]{natbib}

\newcommand{\target}{\theta'_{cs}}
\newcommand{\net}{\theta_{cs}}
\usepackage{amsmath}
\DeclareMathOperator*{\argmax}{arg\,max}

\title{Forgetful Experience Replay in Hierarchical Reinforcement Learning from Demonstrations}

\author{%
Alexey Skrynnik\\
Artificial Intelligence Research Institute FRC CSC RAS\\
Moscow, Russia\\
\texttt{skrynnik@isa.ru} \\
\And
Aleksey Staroverov\\
Moscow Institute of Physics and Technology\\
Moscow, Russia\\
\texttt{alstar8@yandex.ru} \\
\And
Ermek Aitygulov\\
Moscow Institute of Physics and Technology\\
Moscow, Russia\\
\texttt{aytygulov@phystech.edu} \\
\And
Kirill Aksenov\\
Higher School of Economics\\
Moscow, Russia\\
\texttt{kaaksyonov@edu.hse.ru} \\
\And
Vasilii Davydov\\
Moscow Aviation Institute\\
Moscow, Russia\\
\texttt{dexfrost89@gmail.com} \\
\And
Aleksandr I. Panov\\
Artificial Intelligence Research Institute FRC CSC RAS\\
Moscow Institute of Physics and Technology\\
Moscow, Russia\\
\texttt{panov.ai@mipt.ru} \\
}

\begin{document}

\maketitle
\begin{abstract}
Currently, deep reinforcement learning (RL) shows impressive results in complex gaming and robotic environments. Often these results are achieved at the expense of huge computational costs and require an incredible number of episodes of interaction between the agent and the environment. There are two main approaches to improving the sample efficiency of reinforcement learning methods – using hierarchical methods and expert demonstrations. In this paper, we propose a combination of these approaches that allow the agent to use low-quality demonstrations in complex vision-based environments with multiple related goals. Our forgetful experience replay (ForgER) algorithm effectively handles errors in expert data and reduces quality losses when adapting the action space and states representation to the agent's capabilities. Our proposed goal-oriented structuring of replay buffer allows the agent to automatically highlight sub-goals for solving complex hierarchical tasks in demonstrations. Our method is universal and can be integrated into various off-policy methods. It surpasses all known existing state-of-the-art RL methods using expert demonstrations on various model environments. The solution based on our algorithm beats all the solutions for the famous MineRL competition and allows the agent to mine a diamond in the Minecraft environment.
\end{abstract}
\section{Introduction}

Modern methods of reinforcement learning (RL) require huge computational resources and a large number of episodes of interaction with the environment, especially for learning effective policies in complex hierarchical and robotic environments~\cite{2019arXiv191001741Y,2018arXiv180610293K}. 
One of the most promising approaches to developing sample-efficient RL methods is imitation learning~\cite{duan2017one,2018arXiv180209564Z} and the use of expert demonstrations~\cite{DQfD,2017arXiv170708817V}. Expert trajectories for the imitation are obtained by recording human actions or running some pre-trained algorithm. Getting high-quality suboptimal demonstrations is a separate, time-consuming task that is comparable in complexity with data markup for supervised learning \cite{martinez2017relational}. Another challenge is adapting expert trajectories to the capabilities and limitations of agents. It is often impossible to accurately correlate the actions of an expert and the actions of an agent, especially in hybrid discrete-continuous cases.

The standard approach is when we simplify the task of collecting expert data by reducing the quality requirements for trajectories and the optimal strategy used by the expert \cite{gao2018reinforcement,rajeswaran2017learning}. 
In this paper, we consider the use of demonstrations in off-policy RL methods with replay buffer as the most suitable methods for learning and planning on trajectories collected by secondary strategy~\cite{eysenbach2019search,andrychowicz2017hindsight}. Using noisy trajectories, which are applied for initial buffer filling, allows the agent to learn more quickly in the environment. However, ineffective actions in demonstrations are another source of agent errors, in addition to unobserved states that are not covered by expert trajectories. This is even more likely to lead to a catastrophic drop in total reward and it is the main problem in RL from imperfect demonstrations \cite{gao2018reinforcement}.

We offer the new hierarchical experience replay technique that allow us to overcome the main disadvantages of existing methods of off-policy learning from demonstrations. We propose managed forgetting expert data by reducing their ratio in experience replay buffer and learning bathes. In contrast to previously used approaches with a constant priority and a fraction of expert data \cite{PER,DQfD}, this allows us to effectively deal with the influence of low-quality expert actions and effectively adapt trajectories to a discrete space of agent actions. We applied a hierarchical approach to using demonstrations and filling a replay buffer. We proposed an algorithm for extracting sub-goals from expert data that correspond to certain parts of trajectories and subtasks. Following the selected sub-targets, the agent constructs a replay buffer in which parts of the trajectories for each sub-task are explicitly separated. This replay buffer structure allows us to effectively generate meta-actions and tuning them during interaction with the environment. Finally, we developed the hierarchical goal-oriented augmentation of expert data, which allows us to partially reduce the requirement for the amount of data necessary for high-quality imitation.

We used these techniques to develop a new Forgetful Experience Replay (ForgER) algorithm that forging expert data while collecting its own experience automatically correlated with sub-tasks performed in the environment. To investigate the contribution and features of the main ideas, we considered two environment classes – the indicative simple low dimensional environments with simple action space (Lunar Lander, Torcs and others) ~\cite{brockman2016openai} and the complex hierarchical vision-based Minecraft~\cite{johnson2016malmo} with hybrid discrete-continuous action space. On simple environments, we compared ForgER with the well known state-of-the-art (SOTA) approaches and showed the effect of the forgetting technique in three main cases: poor quality of expert data, incorrect choice of discretization, and variability of the environment, which leads to a discrepancy between the expert's value function and the environment in which the agent operates. Unfortunately, most of the known methods that utilise demonstrations are only used in toy environments and they are poorly scalable. In this regard, we have applied our method in the well-known complex Minecraft environment, which has recently served as a good benchmark for testing RL algorithms in rich hierarchical settings~\cite{minerlcomp,shu2017hierarchical}. This allowed us to demonstrate the performance of ForgER and surpass the results of recent SOTA methods in MineRL competition~ \cite{2019arXiv191208664S}.

The main contributions of this work are as follows. We investigate the impact of poor-quality expert data on the effectiveness of off-policy methods. We propose a new forgetting mechanism to deal with a catastrophic drop in productivity due to poor-quality expert trajectories and extend the approach of using demonstrations to partially observed and hierarchical environments. We propose the method of data augmentation in RL that weakens the requirements for the amount and quality of expert data for efficient imitation. We conducted a detailed experimental study on simple environments and were able to show the advantage of our method compared to the known SOTA approaches. Using the proposed method, we surpass the existing approaches to solving hierarchical tasks in the Minecraft environment.
\section{Background}
We consider a Markov decision process (MDP)~\cite{MDP} defined by the tuple $(S,A,T,R,\gamma)$ where $S$ is the state space, $A$ is the action space, the unknown transition probability $T : S \times A \times S \rightarrow [0,\infty)$ represents the probability density of reaching $s_{t+1} \in S$ from $s_t \in S$ by taking the action $a \in A$, $\gamma \in [0, 1]$ is the discount factor, and the bounded real-valued function $R : S \times A \rightarrow [r_{\min}, r_{\max}]$ represents the reward of each transition. 
Using the policy $\pi(a_t|s_t)$ to sequentially generate actions we obtain a trajectory through the environment $\tau = (s_0, a_0, r_0, s_1, a_1, r_1,\dots)$. 
For any given policy, we define the action-value function (Q-function) as $Q^\pi(s, a) = \mathbb{E}_{\tau:s_0=s, a_0=a}[\sum_t\gamma^t R(s_t, a_t)]$ that represents the expected discounted future total return. 
The goal is to learn the optimal policy $\pi^*$, which maximizes the action-value function for every state in $S$~\cite{Sutton}.

Q-learning~\cite{Qlearn} is a well-known RL algorithm that uses samples of experience of the form $(s_t, a_t, r_t, s_{t+1})$ to estimate the optimal action-value function $Q^*(s_t, a_t)$. Hereby, $Q^*(s_t, a_t)$ is the expected return of
selecting action $a_t$ in state $s_t$ and following an optimal policy $\pi^*$. Deep RL methods like DQN~\cite{DQN}, Double DQN~\cite{DoDQN} and Dueling DQN~\cite{DuDQN} parameterize the Q-function and represent it as $Q_\theta(s_t,a_t)$, where neural network weights $\theta$ are updated using stochastic gradient descent. There are two main points in DQN that allowed us to apply deep neural networks to the RL problem. First, it uses a separate target network that is copied every few steps from the regular network. Second, it uses a a replay buffer $\mathcal D$ where the agent adds all of its experiences. The use of these techniques leads to the fact that the target Q-function are more stable.

Most challenging tasks in reinforcement learning provide only partial observations. In our work, we conduct the main experiments in partially observed environments, the model of which is represented as Partially Observable Markov Decision Process (POMDP)~\cite{cassandra1994acting}. A POMDP is a tuple $(S,O,A,T,R,\omega, \gamma)$ where $S, A, R, T$, and $\gamma$ are defined as in an MDP, $O$ is a finite set of observations, and $\omega(s, o)$ is the observation probability distribution. At every time step $t$, the agent executes an action $a_t\in A$, receives reward $R(s_t, a_t)$, and receives an observation $o_{t+1}\in O$. The agent does not observe the true state $s_{t+1}$ and only observation provides the agent a clue about what the state $s_{t+1}\in S$ is. Usually to account for the agent's interaction history with the environment the concept of a belief state if introduced. A belief state is a probability distribution $b_t : S \rightarrow [0, 1]$ over $S$, such that $b_t(s)$ is the probability that the agent is in state $s\in S$ given the history up to time $t$. The belief state $b_{t+1}$ is determined as following $b_{t+1}(s') \propto \omega(s',o_{t+1})\sum_{s\in S}p(s,a_t,s')b_t(s)$ for all $s'\in S$.

\subsection{Q-learning from demonstrations}
The idea of using demonstrations is to learn the agent as much as possible from the demonstration data before running on the real environment. Most of algorithms such as DQfD~\cite{DQfD} work in a fully observable case. The use of demonstrations occurs once - during imitating phase in which the agent learns to imitate the demonstrator. During this imitating phase, the agent samples mini-batches from the demonstration data and updates the network by applying four-component loss function $L(Q)=L_{DQ}(Q)+\lambda_1 L_n(Q)+\lambda_2 L_E(Q)+\lambda_3 L_{L2}(Q)$. There $L_{DQ}(Q)=(R(s_t,a_t)+\gamma Q_{\theta'}(s_{t+1},\argmax_a Q_\theta(s_{t+1},a))-Q_\theta(s_t,a_t))^2$ is the 1-step double Q-learning loss and $L_n(Q)$ is an n-step double Q-learning loss. This part of the overall loss function ensures that the network satisfies the Bellman equation and helps propagate the values of the expert’s trajectory to all the earlier states \cite{A3C}. The main part, $L_E(Q)=\max_{a_t\in A}[Q(s_t,a_t)+l(a_t^E,a_t)]-Q(s_t,a_t^E)$ is a supervised large margin classification loss \cite{Piot2014} where $a_t^E$ is the action the expert demonstrator took in state $s_t$ and $l(a_t^E, a_t)$ is a margin function that is 0 when $a_t = a_t^E$ and positive otherwise. This loss grounds the values of the unseen actions to reasonable values, and makes the greedy policy imitate the demonstrator. At last, $L_{L2}(Q)$ is an L2 regularization loss on the network weights and biases. It helps prevent the network from overfitting on the relatively small demonstration dataset.

When the imitating phase is completed, the agent begins to fill the replay buffer $\mathcal{D}$ with self-generating data that represent its own experience. In such works as PDD DQN~\cite{PER} for forming a sample, prioritization is used. It consists of adding different small positive constants to the priorities of the agent and demonstration transitions to control the relative sampling of demonstration versus agent data.

\section{Forgetful experience replay}

In addition to simple single-goal environments, our approach extends to partially observable hierarchical tasks in which a hierarchical structure of subtasks given or can be extracted. This assumption allows us to simplify the difficult POMDP task with sparse rewards to a set of simpler ones. Each subtask we define as a meta-action or option \cite{sutton1999between}. An option is triple $(\mathcal{I}, \pi, \beta)$ in which $\mathcal{I} \in O$ is the initiation set, $\pi$ is an inner option policy and $\beta$ is a termination function. In addition, each option may have its own function $f_p$ that defines pseudo-rewards. For some tasks, the hierarchical structure can be extracted automatically or semi-automatically using human demonstrations.

We will consider such problems in which subgoals can be determined by the features extracted from expert trajectories. Examples of such features are the appearance of an item in the character’s inventory, reaching a new level, receiving a certain reward.
To represent the dependency of one subtask on another we construct a directed weighted graph $G = (V, E)$, where the vertices $V = \{g_1, g_2, ..., g_n\} $ are the extracted subgoals (feature vectors). The set of edges $E$ is defined by transitions from one subgoal to another in expert trajectories, the weights of the edges is proportional to the number of such transitions. This graph needs to be turned into a tree for applying one of the hierarchical reinforcement learning methods. Topological sorting and various heuristics can be used for this conversion. The resulting tree does not guarantee the convergence of hierarchical approaches, and adjusting its structure on an adhoc basis is a further direction of our work.

Agent learning process consists of two phases - \textit{imitating phase} using demonstrations and \textit{forging phase} when the agent refines policy during interaction with the environment. On both phases the agent uses replay buffer structured regarding the current subgoal tree $G$. During the imitating phase agent samples mini-batches from the demonstration data and updates the network by applying the POMDP goal-specific version of the DQfD loss function $L(Q^g)=L_{PDQ}(Q^g)+\lambda_1 L_n(Q^g)+\lambda_2 L_{PE}(Q^g)+\lambda_3 L_{L2}(Q^g)$ where $g\in V(G)$ is a subtask during which experience is collected, $L_{PDQ}(Q^g)=(R(o_t,a_t)+\gamma Q^g_{\theta'}(o_{t+1},\argmax_a Q^g_\theta(o_{t+1},a))-Q^g_\theta(o_t,a_t))^2$ and $L_{PE}(Q^g)=\max_{a_t\in A}[Q^g(o_t,a_t)+l(a_t^E,a_t)]-Q^g(o_t,a_t^E)$. The structured  replay buffer is defined by $\mathcal{D} = \{d_i=(o,a,r,o',\lambda_2, g)\}$ in which $\lambda_2$ is a margin weight. Each option policy $\pi^g$ is formed using a new hierarchical augmentation approach. During the forging phase, forgetting is used when dynamic adjustment of the ratio of expert and agent experience occurs. In addition to the introduced replay buffer, we are adding noisy layers, which used along with $\epsilon-greedy$ exploration. The pseudocode of the ForgER approach is sketched in Algorithm~\ref{alg:forger}.

\begin{algorithm}[!ht]
    \caption{Forgetful experience replay \label{alg:forger}}
    \begin{algorithmic}[1]
        \STATE Inputs:
        $G$: subtask graph,
        $\theta$: weights for initial subtask network,
        $\theta'$: weights for target subtask network,
        $\tau$: frequency at which to update target net,
        $k$: number of imitating steps,
        $\mathcal{D}$: structured replay buffer,
        $f_p$: goal-oriented pseudo reward function,
        $f_{rg}(t)$: forgetting ratio function
        
        \FOR{$g \in V(G)$}
        \STATE $t_{g} \leftarrow 0$, $\theta_g \leftarrow \theta$, $\theta'_{g} \leftarrow \theta'$, 
        \STATE $\mathcal{D}^{demo}_{g} \leftarrow (o, a, f_p(o, r), o', a', \lambda_2, g):d_i \in \mathcal{D}$ where $\lambda_2=1$,
        \STATE $\mathcal{D}^{extra}_{g} \leftarrow (o, a, 0, o', a', 0, g'):d_i \in \mathcal{D}$ where $\lambda_2=1, g'\neq g$
        \STATE \textbf{imitating}($g, \mathcal D^{demo}_{g}, \mathcal{D}^{extra}_{g}$)
        \ENDFOR

        \FOR{episode $k \in \{ 1, 2, \ldots \}$}
        \STATE \textbf{forging}($g, \mathcal{D}^{demo}_{g}, \mathcal{D}^{agent}_{g}, \ t_{g}, f_p$, $k$, $f_{rg}(k)$)
        \STATE Select next subtask $g$ from $G$
        \ENDFOR
    \end{algorithmic}
\end{algorithm}

\begin{algorithm}[!ht]
    \caption{ForgER forging phase \label{alg:forger:train}}
    \begin{algorithmic}[1]

        \FUNCTION{forging}
        \STATE Inputs:
        $g$: current subtask,
        $\mathcal{D}^{demo}_g$: initialized with demonstration data for current subtask $g$,
        $\mathcal{D}^{agent}_g$: initialized with data collected by agent for current subtask $g$,
        $t$: current step,
        $f_p$: pseudo reward function,
        $k$: current episode,
        $f_{rg}(k)$: forgetting until episode
        \WHILE{subtask $g$ not solved}
        \STATE Sample action from policy $a\sim\pi^{\epsilon Q_{\theta_{g}}}$
        \STATE Play action $a$ and observe $(o', r)$.
        \STATE Replace $(o,a,r,o')$ by $(o,a,f_p(o, r),o')$
        \STATE Store $(o,a,r,o', 0, g)$ in $\mathcal{D}$ 
        \STATE Sample a mini-batch of $n$ transitions from $\mathcal{D}^{demo}_g$ and $\mathcal{D}^{agent}_g$ with forgetting rate $f_{rg}(k)$
        \STATE Calculate loss $L(Q^{g})$ using target network and perform a learning update to $\theta_{g}$
        \STATE {\textbf{if} } $t~\mathbf{mod}~\tau = 0$ {\textbf{then} } $\target \leftarrow \net$ {\textbf{end if} }
        \STATE $t \leftarrow t + 1$
        \ENDWHILE
        \ENDFUNCTION
    \end{algorithmic}
\end{algorithm}

\begin{algorithm}[!ht]
    \caption{ForgER imitating phase \label{alg:forger:pretrain}}
    \begin{algorithmic}[1]
        \FUNCTION{imitating}
        \STATE Inputs:
        $g$: current subtask,
        $\mathcal{D}^{demo}_g$: initialized with demonstration data for current subtask $g$,
        $\mathcal{D}^{extra}_g$: initialized with data from other subtasks $g$

        \FOR{steps $t \in \{ 1, 2, \ldots k\}$}
        \STATE Sample a mini-batch of $n$ transitions from $\mathcal{D}^{demo}_g$ and $\mathcal{D}^{extra}_g$
        \STATE Calculate loss $L(Q^g)$ using target network $\theta'_{g}$ and perform a learning update to $\theta_{g}$
        \STATE {\textbf{if} } $t~\mathbf{mod}~\tau = 0$ {\textbf{then} } $\theta'_{cs} \leftarrow \theta_{g}$ {\textbf{end if} }
        \ENDFOR
        \ENDFUNCTION
    \end{algorithmic}
\end{algorithm}

\subsection{Forgetting in learning from demonstrations}

The forgetting approach is part of our architecture designed for hierarchical tasks, but it can be used separately for learning from the demonstrations tasks, where it showed better results than the standard approach. In this paper, we address these three problems of expert demonstrations, which could be solved using ForgER. The first one is suboptimality of expert's policy, which can be caused by expert errors. The second one is discrepancy between the action space of the agent and expert. For example, data taken from robot sensors may contain noise and errors, imperfect conditions for recording demonstrations, limitation of the space in which the agent can act. The third problem is imbalance of expert trajectories, which may be caused by incorrect data processing (e.g. selection of only the best trajectories with the best initial conditions).

There are several ways to sample a batch from the replay buffer. the first way is sampling in the ratio. Both expert data and agent data were stored in the same buffer. Expert data is always present in the buffer and was sampled with high priority. The second way is sampling occurring in proportion (for example half of the data in batch from the expert, half from the agent)~\cite{R2D3}. In this case, the data of the expert and the agent are stored in separated replay buffers. And the final way is our ForgER approach: data is sampled in dynamic proportion according with $f_{rg}(t)$. The amount of expert data in the batch is gradually decreasing.

Forgetting is the process of dynamically changing the sampling rate of experts and agent data. For example, we can define the sampling rate changing process as $f_{rg}(k) = \min(1, k/d)$ (linear forgetting) in which $f_{rg}(k)$ is a sampling rate (forgetting rate), $k$ is a current episode and $d$ is the last episode of forgetting (i.e. last episode in which expert data is used for learning) (see Algorithm~\ref{alg:forger:train}).

\subsection{Task-specific augmentation}

The idea of hierarchical augmentation is to use data from other subtasks as extra data on the imitating phase for each policy (see Algorithm~\ref{alg:forger:pretrain}). Both supervised loss function $L_{PE}$ and pseudo rewards $f_p$ is turned off for the extra data.

Using this type of augmentation of ForgER we can solve two problems. Margin loss function causes the agent to learn how to act as an expert at the cost of generalization. Additional data prevents overfitting. The division into subtasks leads to the fact that only part of the data is used to learn each option policy $\pi$. The use of additional data and TD losses allows us to learn the agent on additional information from other subtasks. For example, in Minecraft, such behavior as avoiding obstacles or floating out of the water can be reused in different subtasks.

\begin{figure}[htp]
    \centering
    \includegraphics[width=\linewidth]{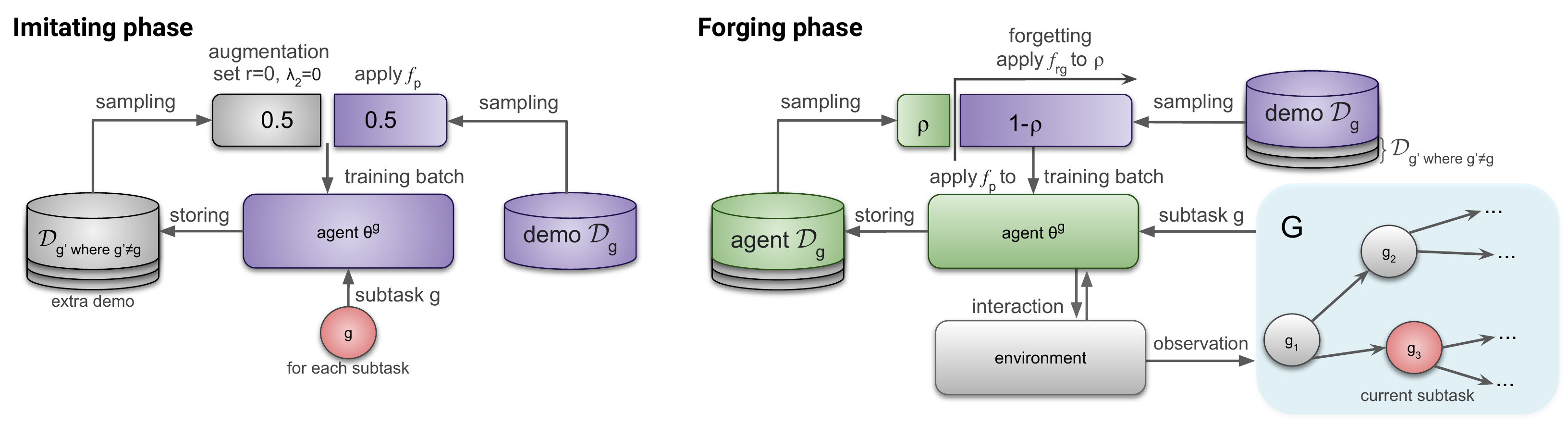}
    \caption{Full ForgER architecture in hierarchical setting.}
    \label{fig:forgfull}
\end{figure}

\begin{figure}[htp]
    \centering
    \includegraphics[width=8cm]{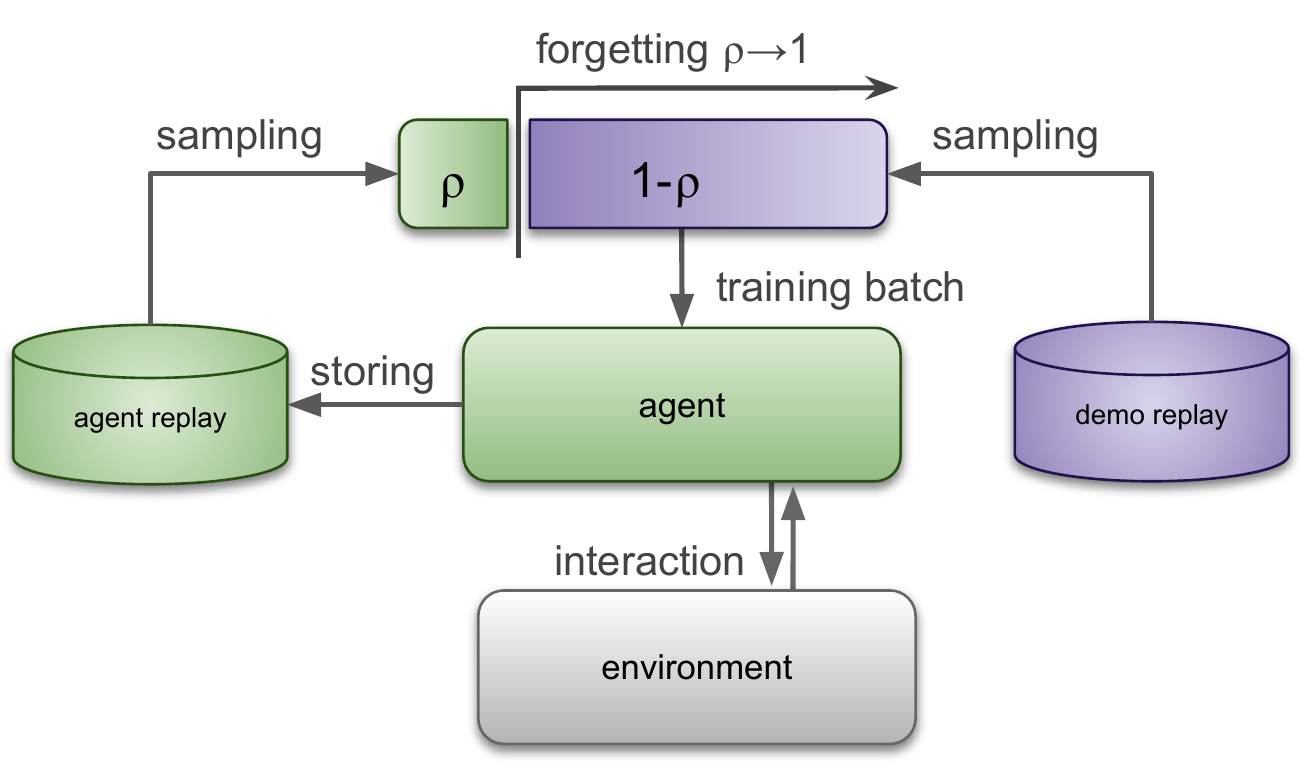}
    \caption{Simplified ForgER architecture. Forgetting is a universal approach for learning from demonstration tasks. Sampling from the expert and agent buffers occurs with a certain ratio function $\rho=f_{rg}(t)$ (forgetting rate), which gradually reduces the amount of used expert data.}
    \label{fig:forgetting}
\end{figure}

The architecture of the agent with the forgetful experience replay buffer is shown in Figure~\ref{fig:forgetting} and Figure~\ref{fig:forgfull}.

\section{Experiments}
We evaluated ForgER in the two classes of environments: \textit{simple} and \textit{hierarchical} sets. The first \textit{simple set} is vector-based and small vision-based environments with a single subgoal: classic Control benchmark (CartPole, MountainCar, Acrobot), environments from Box2D benchmark (LunarLander, CarRacing) \cite{brockman2016openai} and racing simulator \textit{Torcs} with discrete action space. Using this set of environments we compared our method with state-of-the-art methods dealing with expert demonstrations: POfD~\cite{kang2018policy}, NAC~\cite{gao2018reinforcement} and DQfD algorithms. 

The second \textit{hierarchical set} is vision-based \textit{Minecraft} environments including the setting of the \textit{MineRL} competition~\cite{minerlcomp} with hybrid action space. Using this set we demonstrated the behavior of the ForgER agent in a hierarchical setting where different sub-goals can appear when the main goal is reached: \textit{Navigation} (agent must reach the target), \textit{Treechop} (the agent needs to chop wood starting with an iron axe for cutting trees), \textit{ObtainIronPickaxe} (obtaining an iron pickaxe extracting many items sequentially), \textit{ObtainDiamond} (obtaining a diamond, which is the rarest element in Minecraft). The \textit{hierarchical set} is utterly complex since the environment for each of the subtasks is procedurally generated. Also environments from this set have sparse rewards (even dense versions) and belong to a POMDP class because only the first-person view is available for the agent.

In addition to the comparative analysis, we conducted an ablution study to analyze the impact of the quality of expert data and various hyper parameters of our algorithm.

\subsection{MineRL environments}

MineRL is a shell on the game Minecraft, which presents several environments (subtasks):
\begin{itemize}
    \item \textit{Navigate}: In this environment, agent must reach the target. In addition to standard observations, agent has access to a compass that points to a set target located 64 meters from the starting location. Agent receives reward +100 for achieving the goal, after which the episode ends. There is also a Dense version in which the agent receives a reward for each step, depending on the approach to the goal.
    \item \textit{Treechop}: In this environment, the agent needs to chop wood. The agent starts in a forest with an iron ax for cutting trees. The agent receives +1 for each unit of wood received, and the episode ends as soon as the agent receives 64 units or a time limit is reached.
    \item \textit{ObtainIronPickaxe}: Main goal of this environment is an iron pickaxe. To solve this environment, it is necessary to extract many items, and this must be done sequentially. That is, for the extraction of an iron pickaxe, it is necessary to adhere to a hierarchy. There are also two versions of this environment: in the first, the reward is received for the item obtained for the first time, and in the Dense environment for each receipt.
    \item \textit{ObtainDiamond}: Main goal of this environment is a diamond, which is the rarest element in Minecraft. This environment is similar to the previous one, however, after receiving the iron pickaxe, the game does not stop and with its help it is also necessary to get a diamond. As for \textit{ObtainIronPickaxe}, there are 2 versions: regular one and Dense.
\end{itemize}

As mentioned above for environments \textit{ObtainIronPickaxe} (except diamond) and \textit{ObtainDiamond}, rewards are given to the agent only for receiving item:

 \begin{table}[H]
    \label{tabular:reward_diamond}
    \begin{center}
        \begin{tabular}{|c|c|}
            \hline
            \textbf{item} & \textbf{reward} \\ 
            \hline
            log & 1 \\
            \hline
            planks & 2 \\
            \hline
            stick & 4 \\
            \hline
            crafting table & 4 \\
            \hline
            wooden pickaxe & 8 \\
            \hline
            cobblestone & 16 \\
            \hline
            furnace & 32 \\
            \hline
            stone pickaxe & 32 \\
            \hline
            iron ore & 64 \\
            \hline
            iron ingot & 128 \\
            \hline
            iron pickaxe & 256 \\
            \hline
            diamond & 1024 \\
            \hline
        \end{tabular}
    \end{center}
    \caption{Rewards for \textit{ObtainDiamond} and \textit{ObtainIronPickaxe}.}
 \end{table}

It is also worth mentioning that, as observations, the agent receives an colored image with resolution $(64, 64)$ for all environments, also \textit{ObtainDiamond} and \textit{ObtainIronPickaxe} receive inventory dictionary, in \textit{Navigate} environment value of the compass indicating the target is also obtained. But the space of actions is much more complicated and seems to be hybrid, that is, in addition to discrete values, there are also continuous ones.

 \begin{table}[H]
    \label{tabular:actionspace}
    \begin{center}
        \begin{tabular}{|c|c|}
            \hline
            \textbf{action} & \textbf{type} \\
            \hline
            attack & Discrete\\
            \hline
            back & Discrete\\
            \hline
            camera & Box(2)\\
            \hline
            craft & Enumerated\\
            \hline
            equip & Enumerated\\
            \hline
            forward & Discrete\\
            \hline
            jump & Discrete\\
            \hline
            left & Discrete\\
            \hline
            nearbyCraft & Enumerated\\
            \hline
            nearbySmelt & Enumerated\\
            \hline
            place & Enumerated\\
            \hline
            right & Discrete\\
            \hline
            sneak & Discrete\\
            \hline
            sprint & Discrete\\
            \hline
        \end{tabular}
    \end{center}
    \caption{Actions in MineRL environments.}
 \end{table}

However \textit{nearbyCraft}, \textit{nearbySmelt}, \textit{craft} and \textit{equip} actions can be used only in \textit{ObtainDiamond} and \textit{ObtainIronPickaxe} environments.

\subsection{Comparative analysis}
\begin{figure*}[ht]
    \centering
    \subfigure{\includegraphics[width=0.31\linewidth]{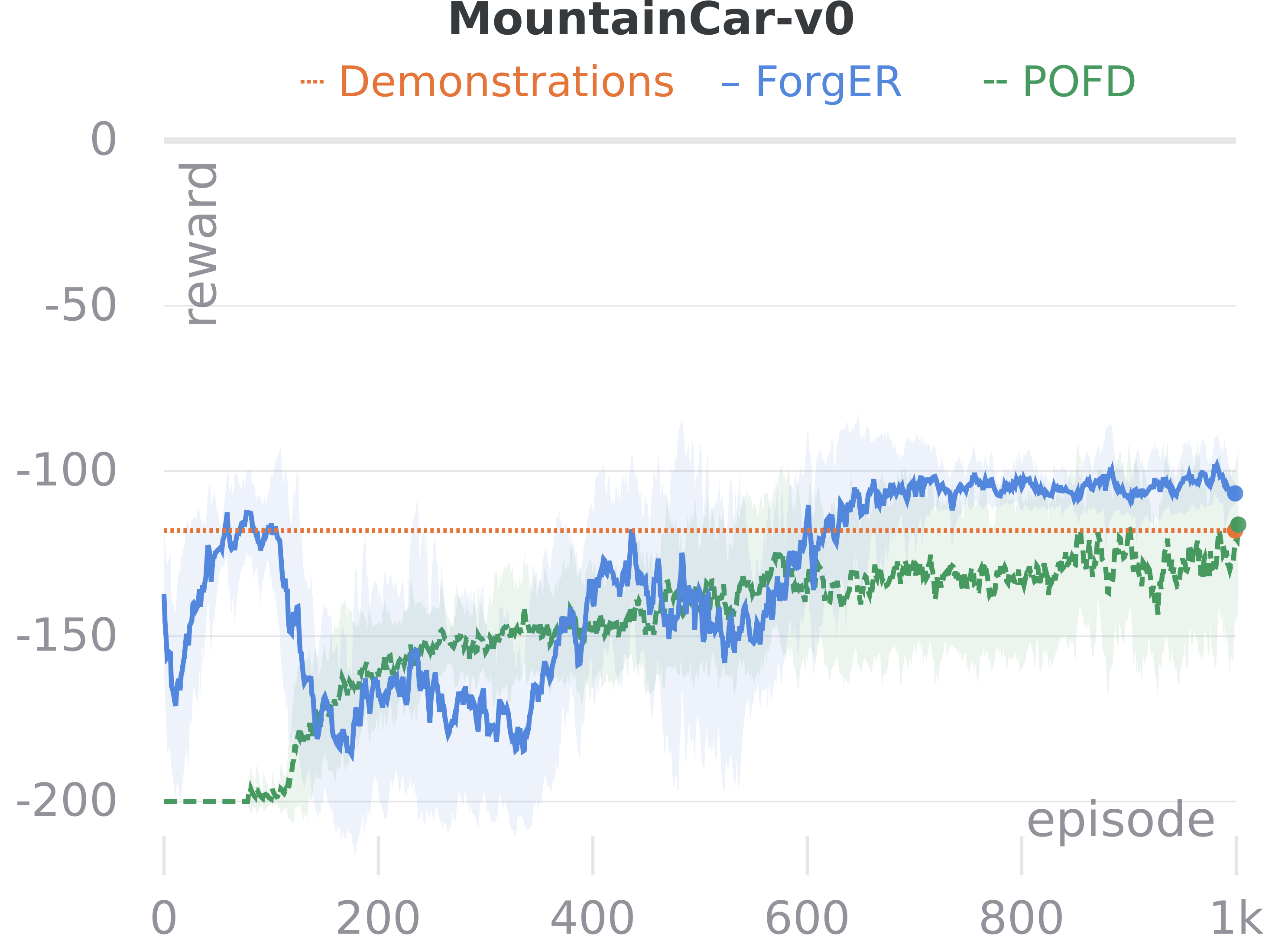}}
    \subfigure{\includegraphics[width=0.31\linewidth]{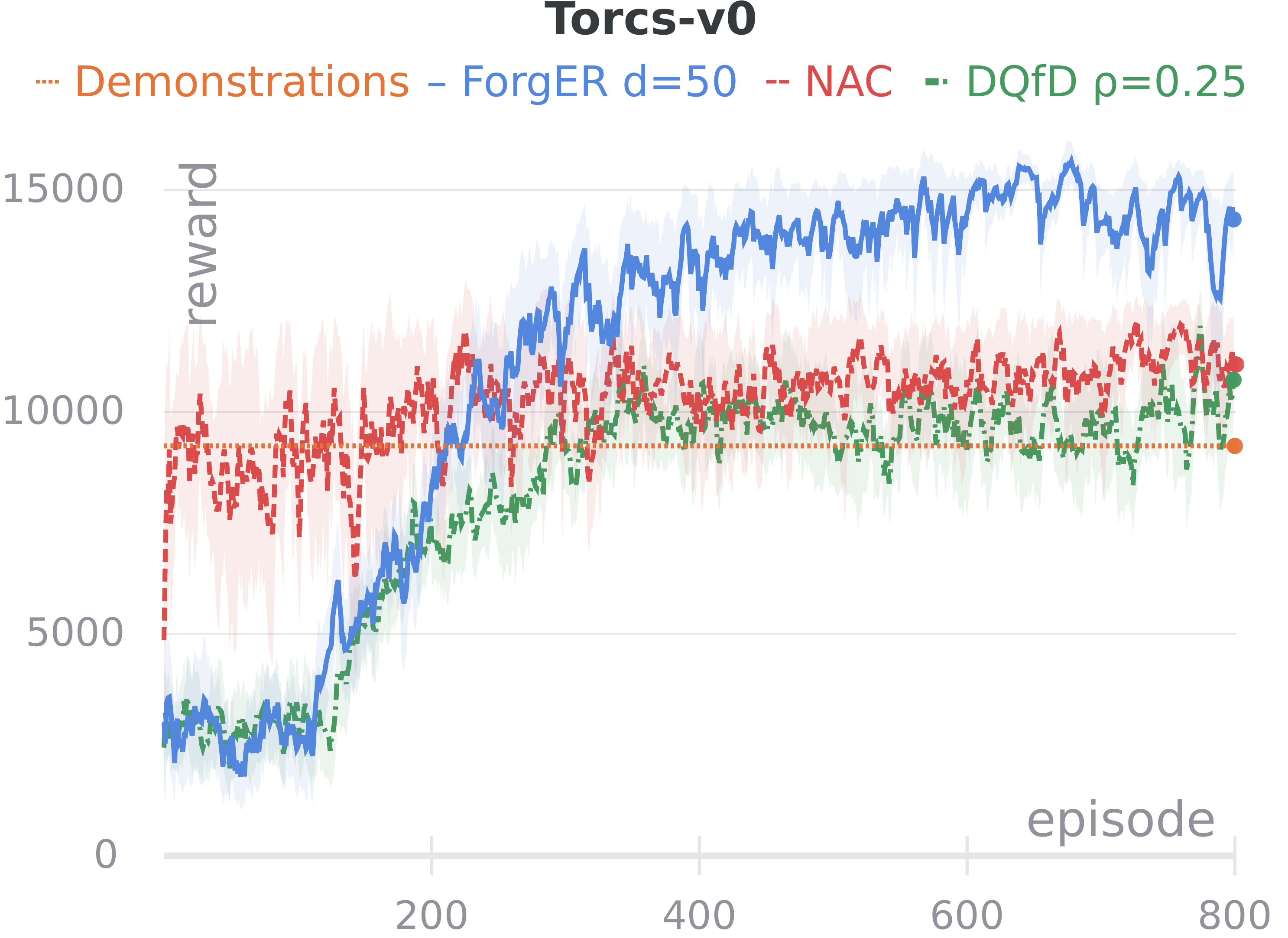}}
    \subfigure{\includegraphics[width=0.33\linewidth]{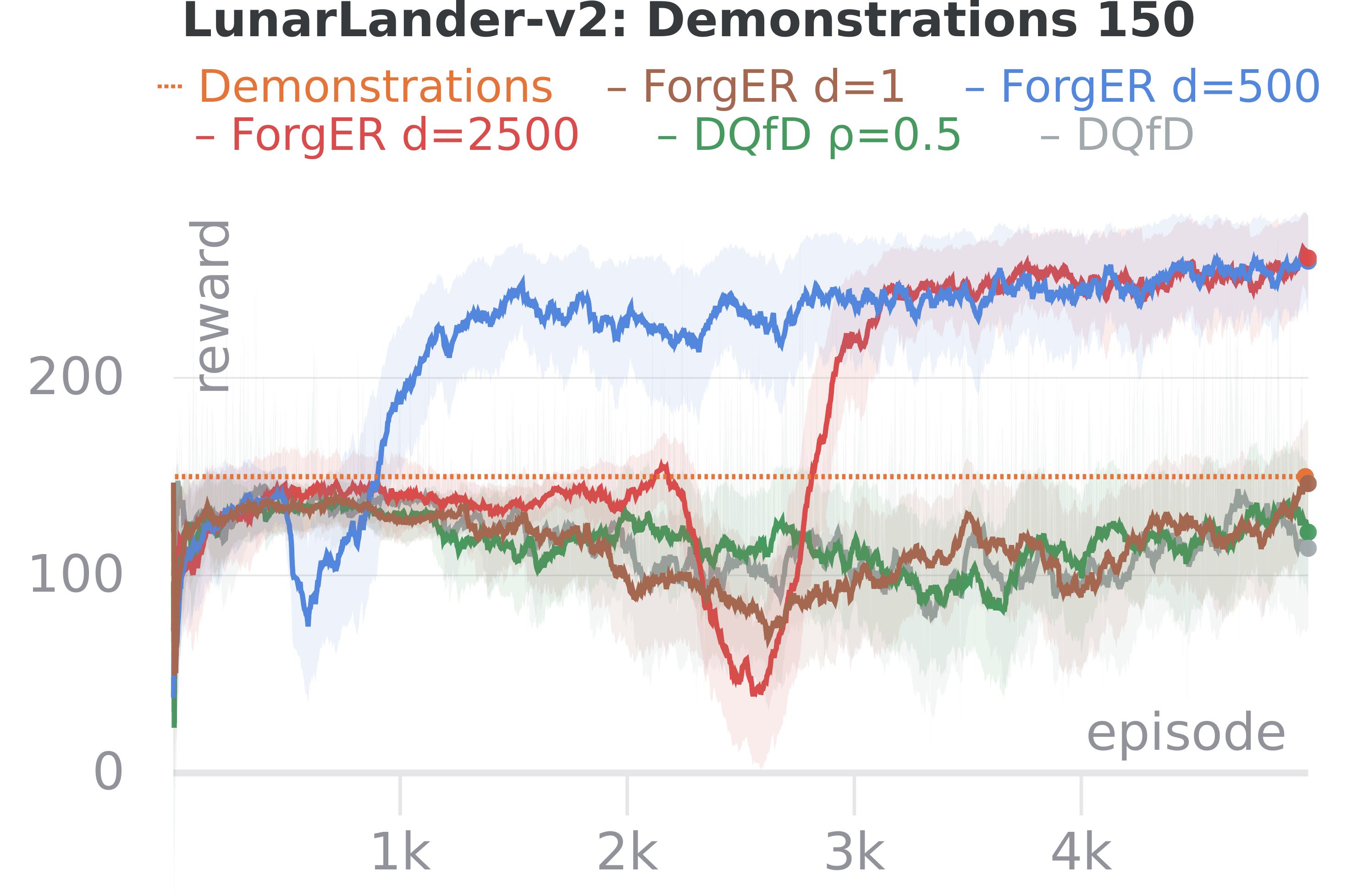}}
    \vspace{-0.2cm}
    \caption{(\textbf{left}) Mean episode reward with high-level expert data for PofD and ForgER in \textit{MountainCar}. (\textbf{middle}) Mean episode reward for FordER, NAC, DQfD agents in \textit{Torcs}. (\textbf{right}) Mean episode reward with high-level expert data for ForgER and DQfD in \textit{LunarLander}.}
    \label{fig-compar}
\end{figure*}

\textbf{ForgER vs POfD in \textit{simple set}}. This set of experiments shows if ForgER performance with imitation phase on demonstrations collected with suboptimal policies can be compared with POfD. POfD was chosen as baseline, because it purposed on getting benefits from imperfect demonstration and outperformed strong algorithms on environments with vector-based observation space. We were able to show that with comparable results obtained with good quality of demonstrations, we significantly outperform POfD when the quality of expert data decreases, including due to changes in the discretization of the action space (an example in Figure~\ref{fig-compar} left).

We evaluated ForgER on three environments that have discrete action space and for each environment we collected 2 set of demonstrations with different average score. ForgER showed comparable performance on demonstrations with high average score, but was outperformed by POfD on demonstrations with low average score. However usually low quality of demonstrations is caused not by bad experts performance. Normally demonstrations are collected with high average score but in setup different from the one in which agent is acting. For example demonstrations can be collected by humans with continuous "camera" actions which were discretized for the agent. For this reason, Box2D benchmark was used. We collected demonstrations with high average score in environments with continuous action space (\textit{LunarLanderContinuous}, \textit{CarRacing}) and then actions where discretized. We used \textit{LunarLander} action space (4 actions) for \textit{LunarLanderContinuous} discretization and custom discretization for \textit{CarRacing} (4 actions). Action from continuous action space was mapped to nearest action from discretized space. In this case ForgER fully outperformed POfD (see Figures 4, 5, 6 and 7). For each algorithm and for each (demonstration, environment) pair only one hyperparameter was tuned: forgetting speed for ForgER ($d$ parameter in linear forgetting function $f_rg(k)$) and reward coefficient for POfD. Results were averaged across 4 seeds. Minimum demonstration proportion was set to 0.0 even in case with good demonstrations to show harmful effect on policy.

\begin{figure*}[ht]
    \centering
    \subfigure{\includegraphics[width=0.48\linewidth]{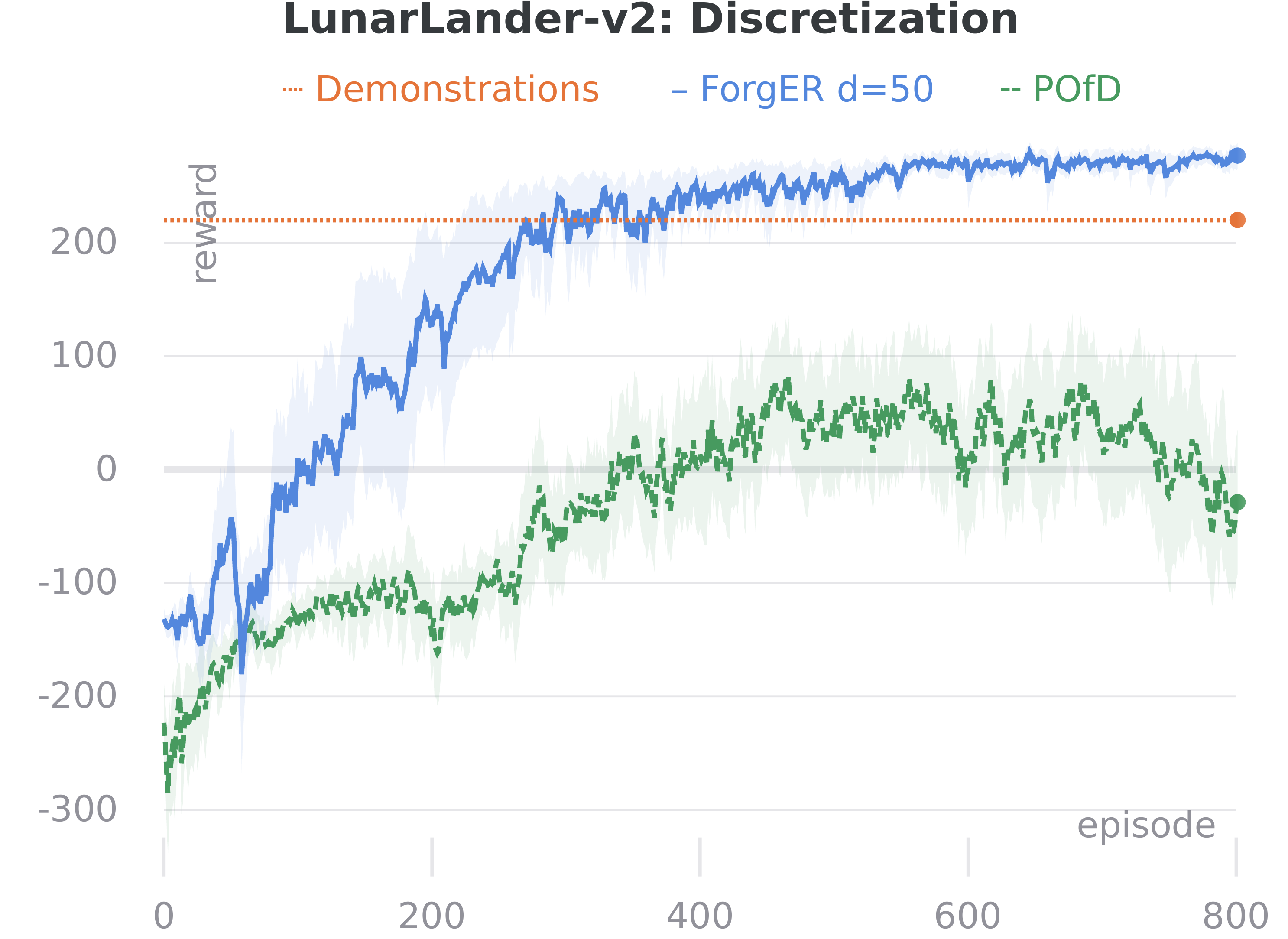}}
    \hspace{0.02\linewidth}
    \subfigure{\includegraphics[width=0.48\linewidth]{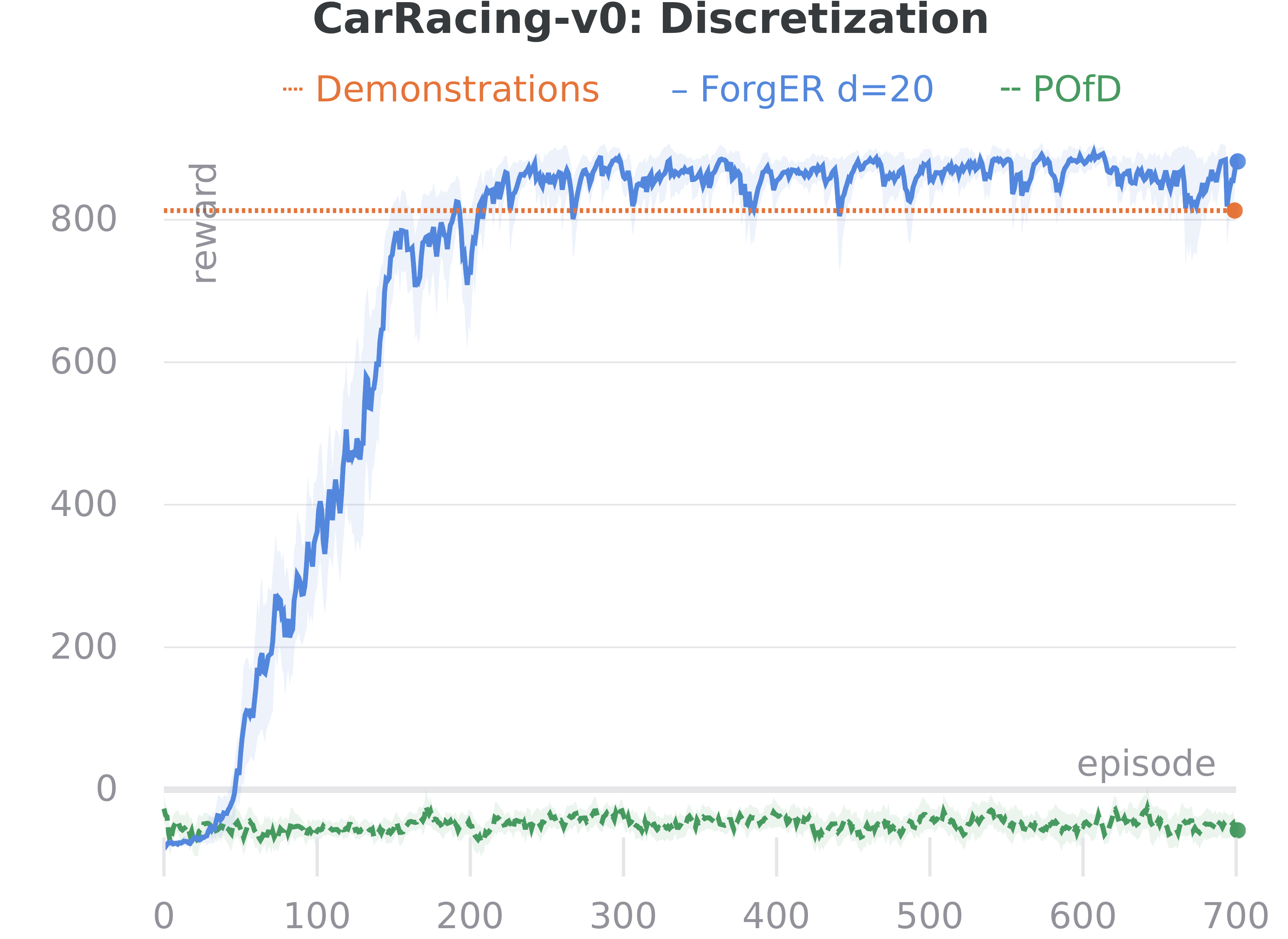}}

    \vspace{-0.2cm}
    \label{fig:exp:discrete:pofd}
    \caption{ForgER vs POfD in \textit{LunarLander} (\textbf{left}) and \textit{CarRacing} (\textbf{right})  with expert data after discretization.}
\end{figure*}

\begin{figure*}[ht]
    \centering
    \subfigure{\includegraphics[width=0.48\linewidth]{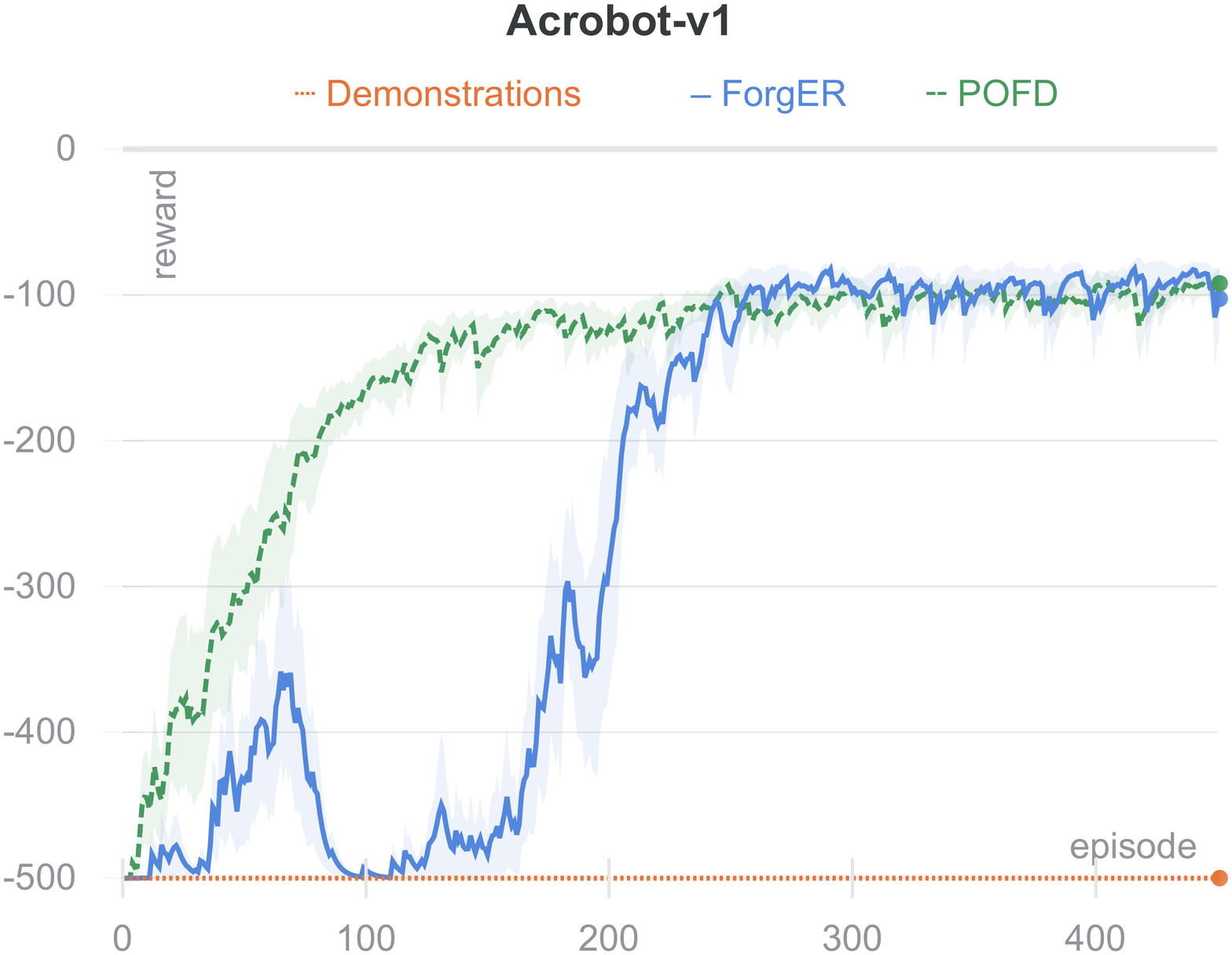}}
    \hspace{0.02\linewidth}
    \subfigure{\includegraphics[width=0.48\linewidth]{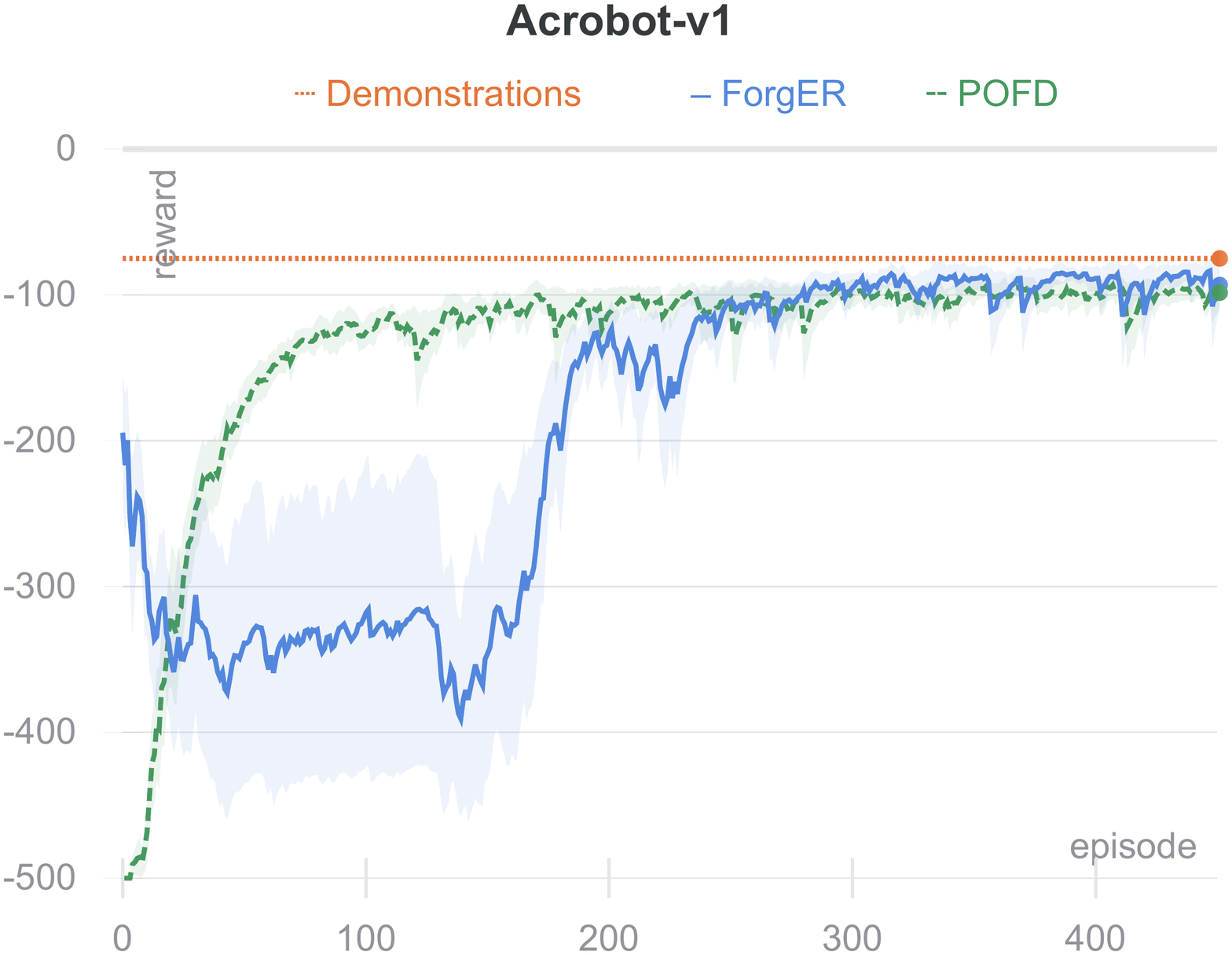}}
    \vspace{-0.2cm}
    \label{fig:exp:abot}
    \caption{Mean episode reward for ForgER (blue) and  POfD (green) agents in \textit{Acrobot} environment and using demonstrations with different average score (orange).}
\end{figure*}

\begin{figure*}[ht]
    \centering
    \subfigure{\includegraphics[width=0.48\linewidth]{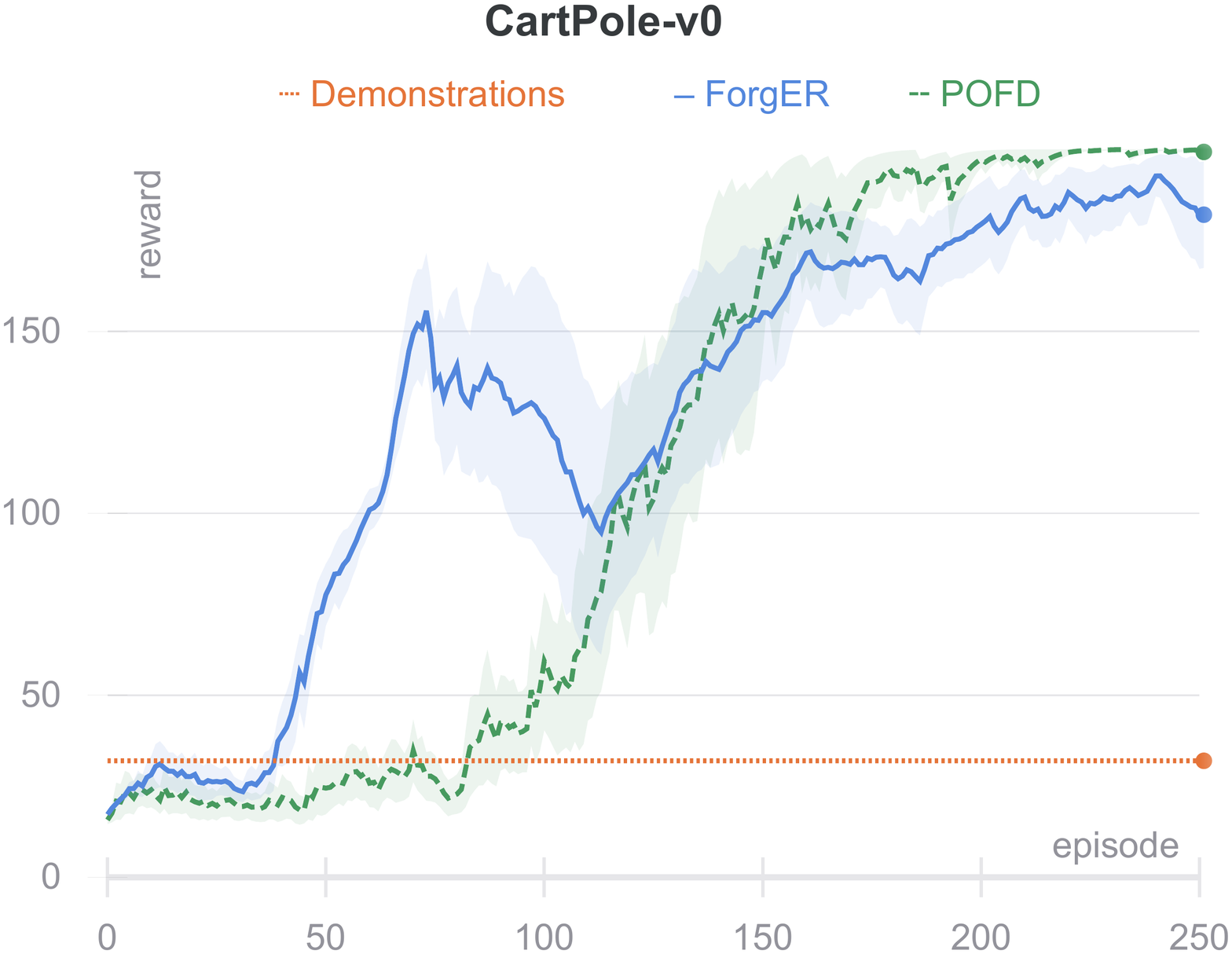}}
    \hspace{0.02\linewidth}
    \subfigure{\includegraphics[width=0.48\linewidth]{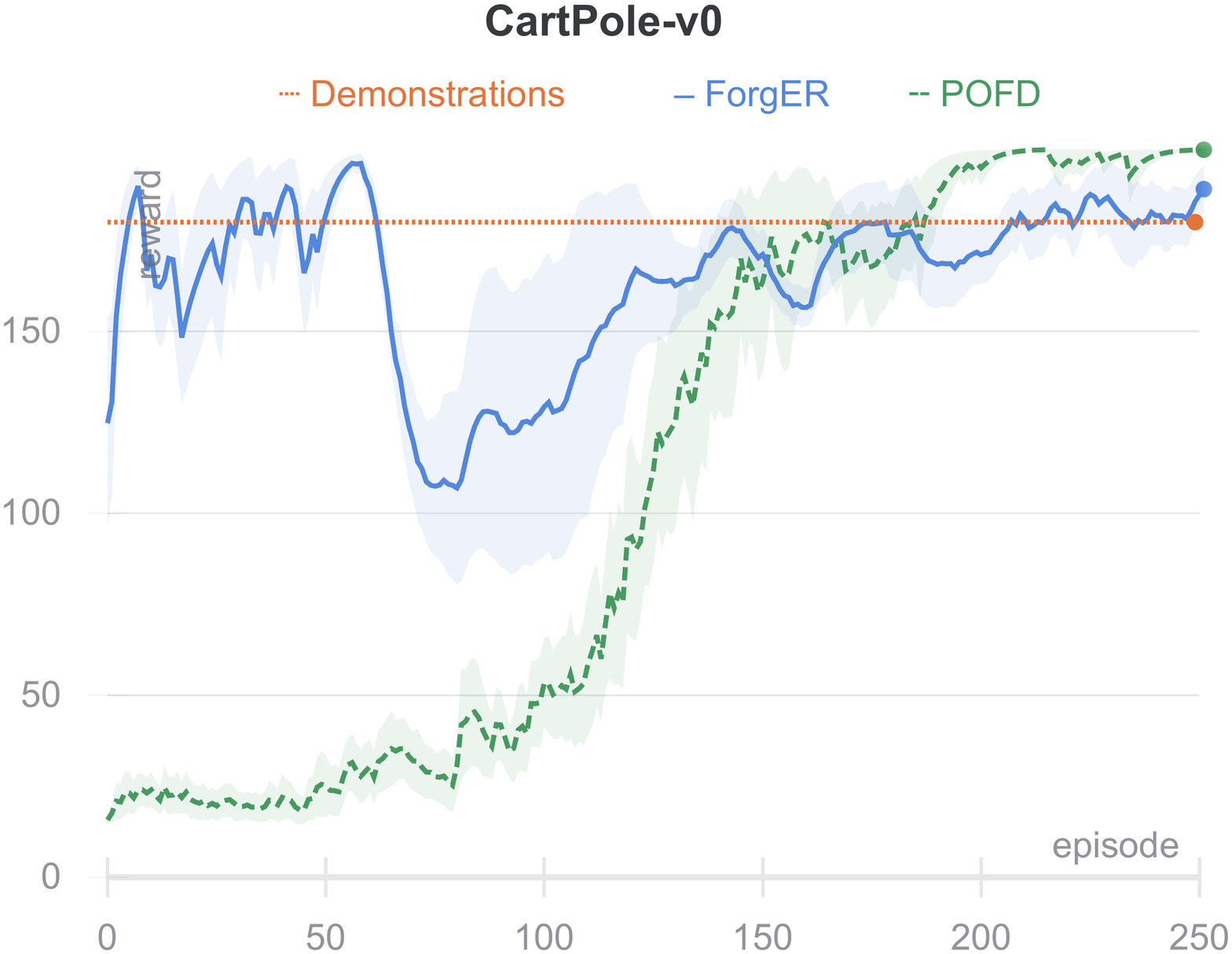}}
    \vspace{-0.2cm}
    \label{fig:exp:cpole}
    \caption{Mean episode reward for ForgER (blue) and  POfD (green) agents in \textit{CartPole} environment and using demonstrations with different average score (orange).}
\end{figure*}

\begin{figure*}[ht]
    \centering
    \subfigure{\includegraphics[width=0.48\linewidth]{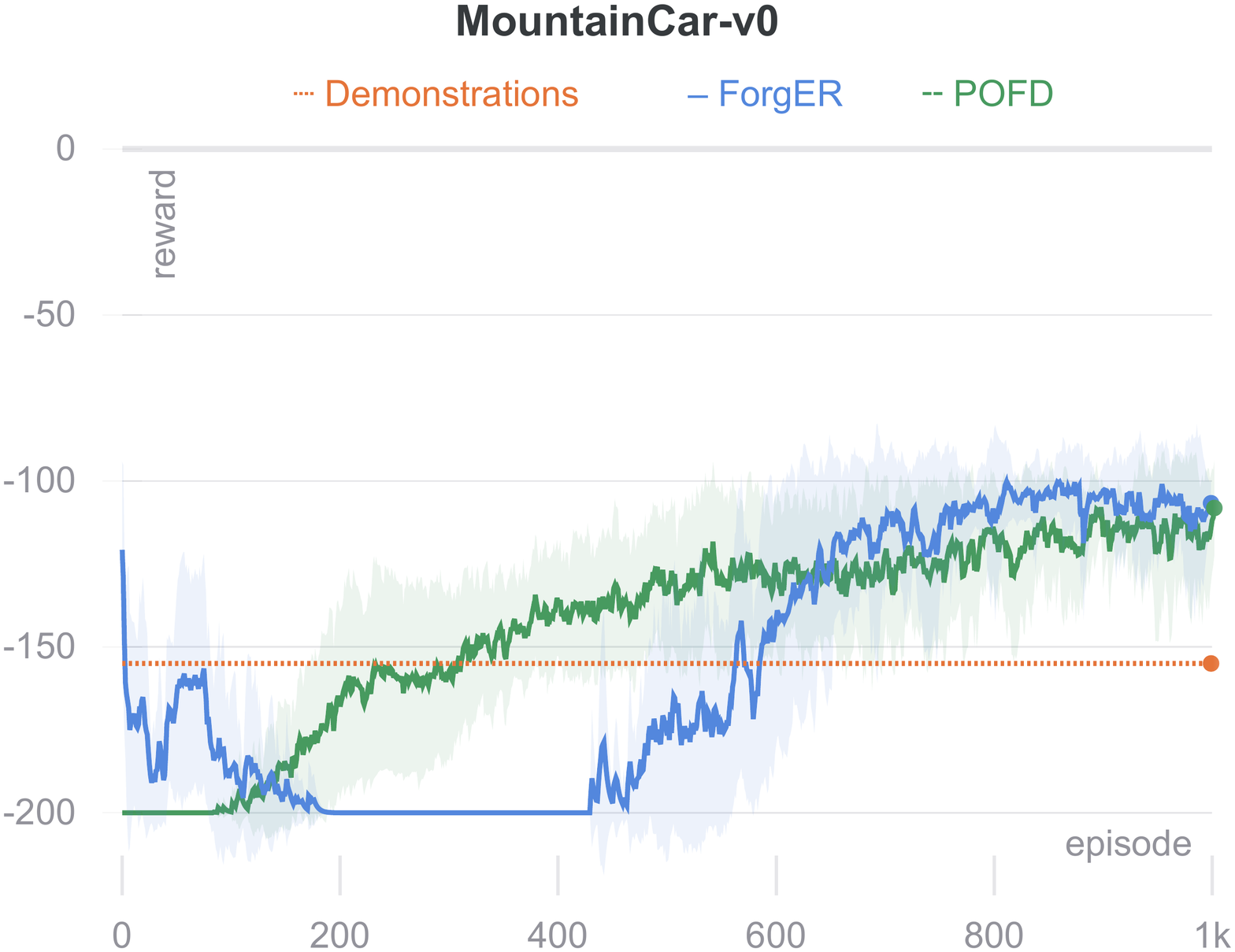}}
    \hspace{0.02\linewidth}
    \subfigure{\includegraphics[width=0.48\linewidth]{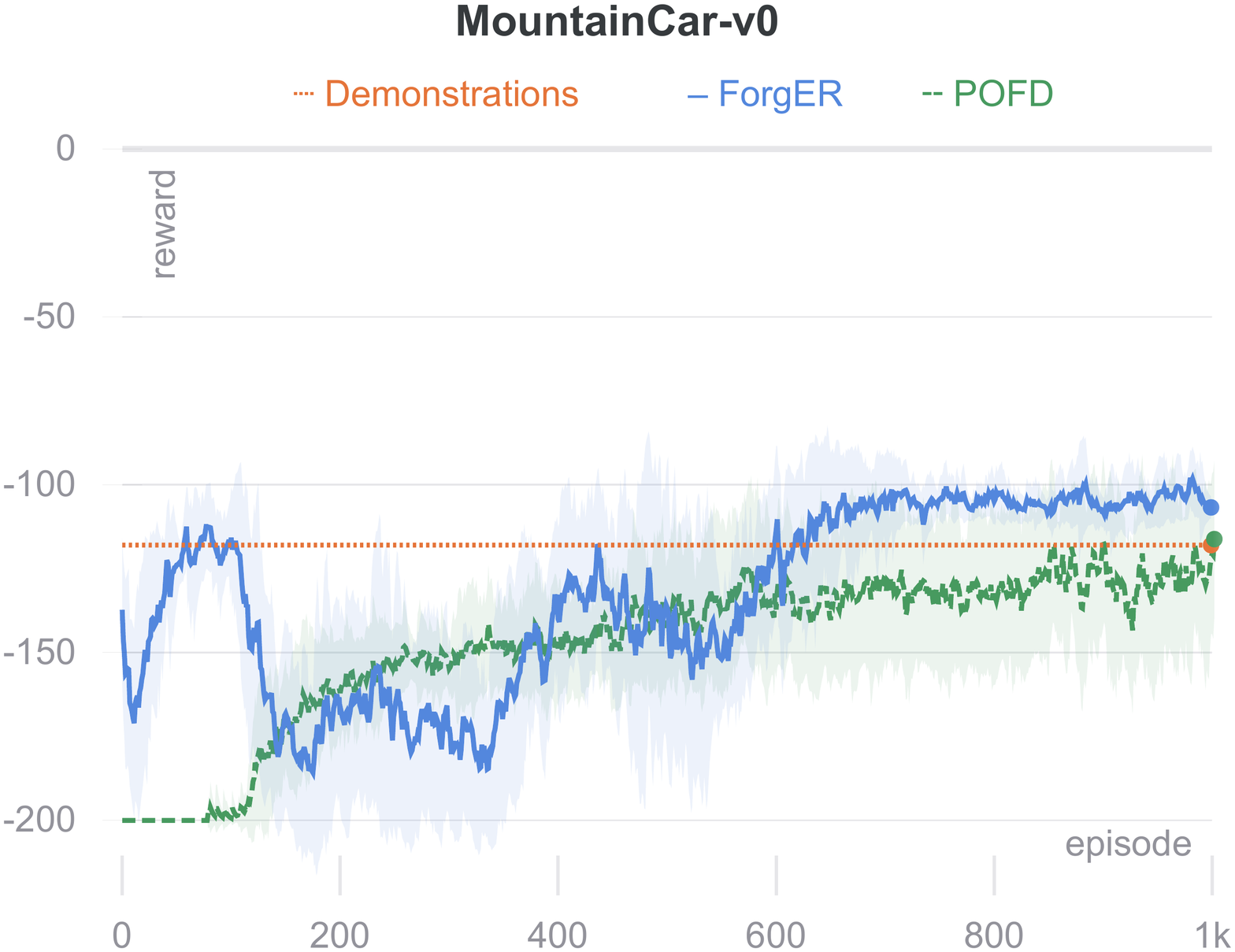}}
    \vspace{-0.2cm}
    \label{fig:exp:mcar}
    \caption{Mean episode reward for ForgER (blue) and  POfD (green) agents in \textit{MountainCar} environment and using demonstrations with different average score (orange).}
\end{figure*}

\textbf{ForgER vs NAC vs DQfD in \textit{simple set}}. In Torcs the goal of the agent is to drive as fast as possible. Action space is a Cartesian product between {left, no-op, right} and {up, no-op, down}, while observation space is vector with size of 29, in which contained all information about the car and the track. Reward is computed at each step and it depends on the velocity of the car projected along the track direction. Expert dataset was obtained using a PPO agent and has an average reward of 9230. In ForgER a linear function with $d=50$ was used as the forgetting function $f_rg(k)$. Results were averaged over 5 random seeds. As can be seen on Figure~\ref{fig-compar} (middle) results of NAC and DQfD are similar to those obtained in the article \cite{gao2018reinforcement}. However, ForgER results are much higher (about $20\%$ compared to NAC).

\subsection{Ablution study}
In this series of experiments, our goal was to demonstrate the impact of expert data quality, action discretization options, task-oriented augmentation on the behavior of ForgER and DQfD, and characteristics of the overfitting process.

\textbf{Impact of expert data quality}. To explore the impact of quality of expert data on the efficiency of off-policy methods we use a well-known environment \textit{Lunar Lander}. Our experiment consists of three parts with different levels of expert data quality. In the first one, high-level expert trajectories with total rewards from 100 to 200 were taken. The second one contains medium-level trajectories with episode rewards from 0 to 100. The third one contains low-level trajectories with lunar lander being broken with episode rewards from -100 to 0. All of these trajectories were taken from one pre-trained agent, picking different trajectories for each part of the experiment. Also in each part, we tested four cases: with constant demo-ratio 0.5, with fully forget after imitation stage, with forgetting until episode 500, with forgetting until episode 2500 and comparison of these results with DQfD algorithm and original pre-trained policy. In the second and third cases forgetting rate changed linearly during the training phase. After imitating phase we sample 100\% of a batch from the expert data buffer. And this rate decreases to 0\% to the episode, after which we want to fully forget expert trajectories. The purpose of these experiments is to prove that forgetting expert data regardless of their quality is improving the learning process and forgetting rate is crucial in this process. All experiment curves are a mean of 10 separate experiment with same parameters.

With high quality expert data (see Figures 2 (right), 8 and 9), by changing the forgetting rate and allocating agent and expert data in replay buffer we can achieve a higher reward. In linearly forgetting cases we can see that significant improvement appears only after we stop using expert data. When the fraction of expert data becomes low, for a moment performance of an agent drops, but then immediately grows up and overcomes expert data by far. The longer we train on expert data, the wider this period is. That happens because when we maintain expert data, we maintain expert policy, that is not optimal and getting out of this local minimum takes more time. But this suboptimal policy guides the agent in the right direction in the early stage, so as we can see that forgetting expert data too early is not giving a performance at all. Forgetting rate is a hyperparameter that needs to be configured for each task.

It would seem that with worse expert data, the performance also should be worse, but that's not always true. In fact, the worse replay buffer includes of more diverse state and action pairs, so the expert policy is not so precise and margin loss $L_{PE}(Q^g)$ can not overfit it. It's like a rookie trying to act like a pro and use advance skills, the chances that one fails while trying to perform this is higher than just using mid-level skills with more awareness. That's a case in the vanilla DQfD approach since our approach deals with the right forgetting rate and outperforms others demo-rate approaches.

\begin{figure}[htp]
    \centering
    \includegraphics[width=8cm]{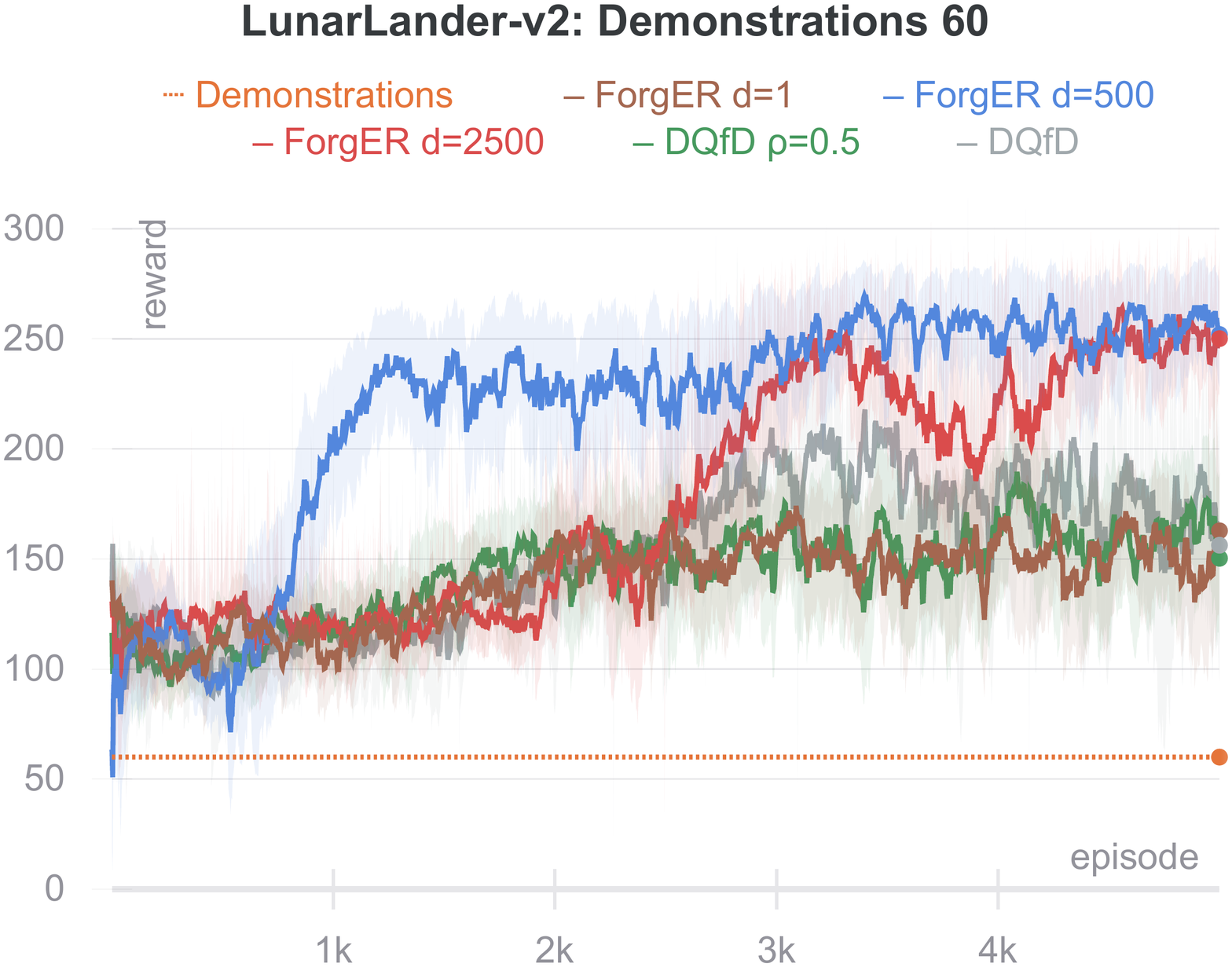}
    \caption{Mean episode reward with medium-level expert data. Due to the wider coverage of states during expert trajectories, the vanilla DQfD approach draws more out of it and gets more reward, but linear forgetting still outperforms it.}
    \label{fig:lunar1}
\end{figure}

\begin{figure}[htp]
    \centering
    \includegraphics[width=8cm]{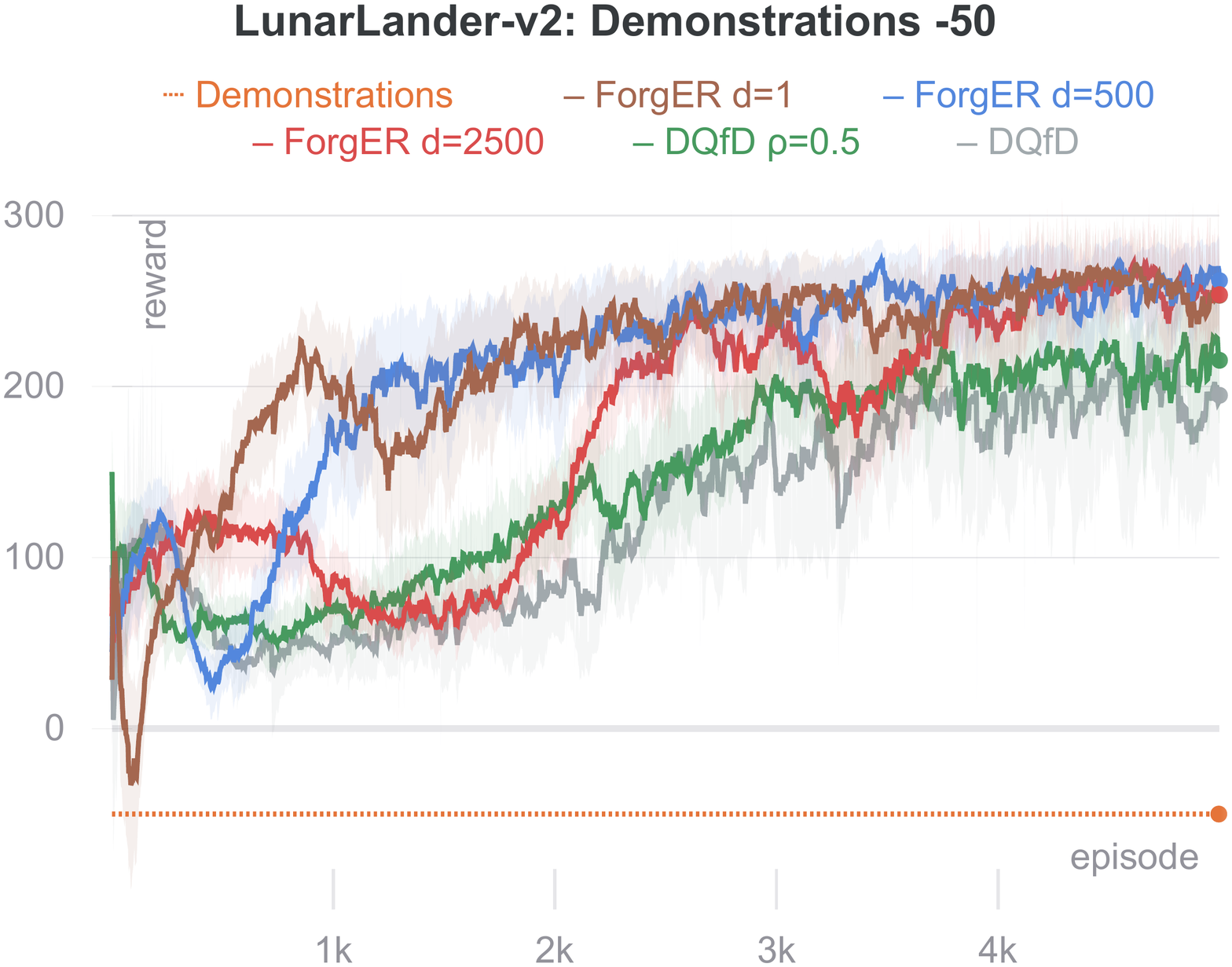}
    \caption{Mean episode reward with low-level expert data. This data contains no useful policy at all so the faster we forget it, the higher our reward will be.}
    \label{fig:lunar2}
\end{figure}

\textbf{Overfitting}. In this experiment, we consider the problem of overfitting on the demonstrations in \textit{Treechop} environment, which caused in the use of margin loss $L_{PE}(Q^g)$. We created a salience maps for imitation (agent after imitating phase, same for DQfD and ForgER), DQfD and ForgER agents. The ForgER agent focusing on important details: crowns and trunks of trees. This fact of overfitting is confirmed by the TD loss diagram on Figure~10 (right). TD loss is strongly correlated with the $f_{rg}$ grow parameter and behaves differently after imitating phase on Figure~10 (left) in comparison to DQfD. TD loss of the ForgER agent after the forging phase is almost 2 times less than the DQfD one. The ForgER agent shows better performance: 62 vs 49 for the \textit{Treechop} task.

\begin{figure*}[ht]
    \centering
    \subfigure{\includegraphics[width=0.6\linewidth]{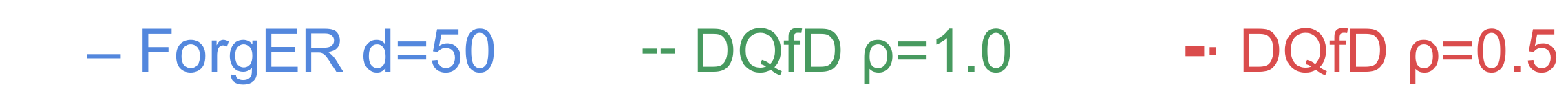}}\\
    \vspace{-0.3cm}
    \subfigure{\includegraphics[width=0.3\linewidth]{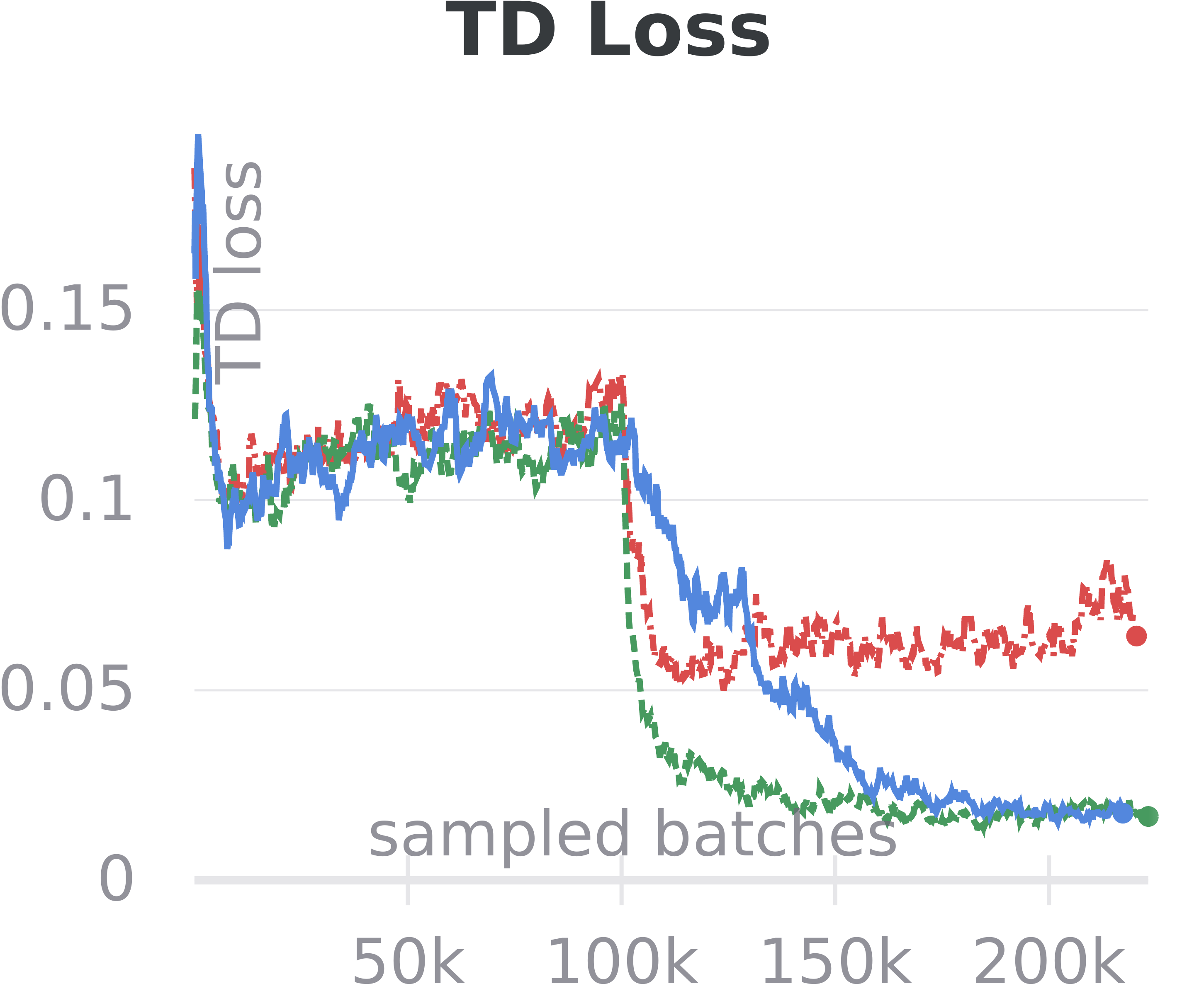}}
    \subfigure{\includegraphics[width=0.3\linewidth]{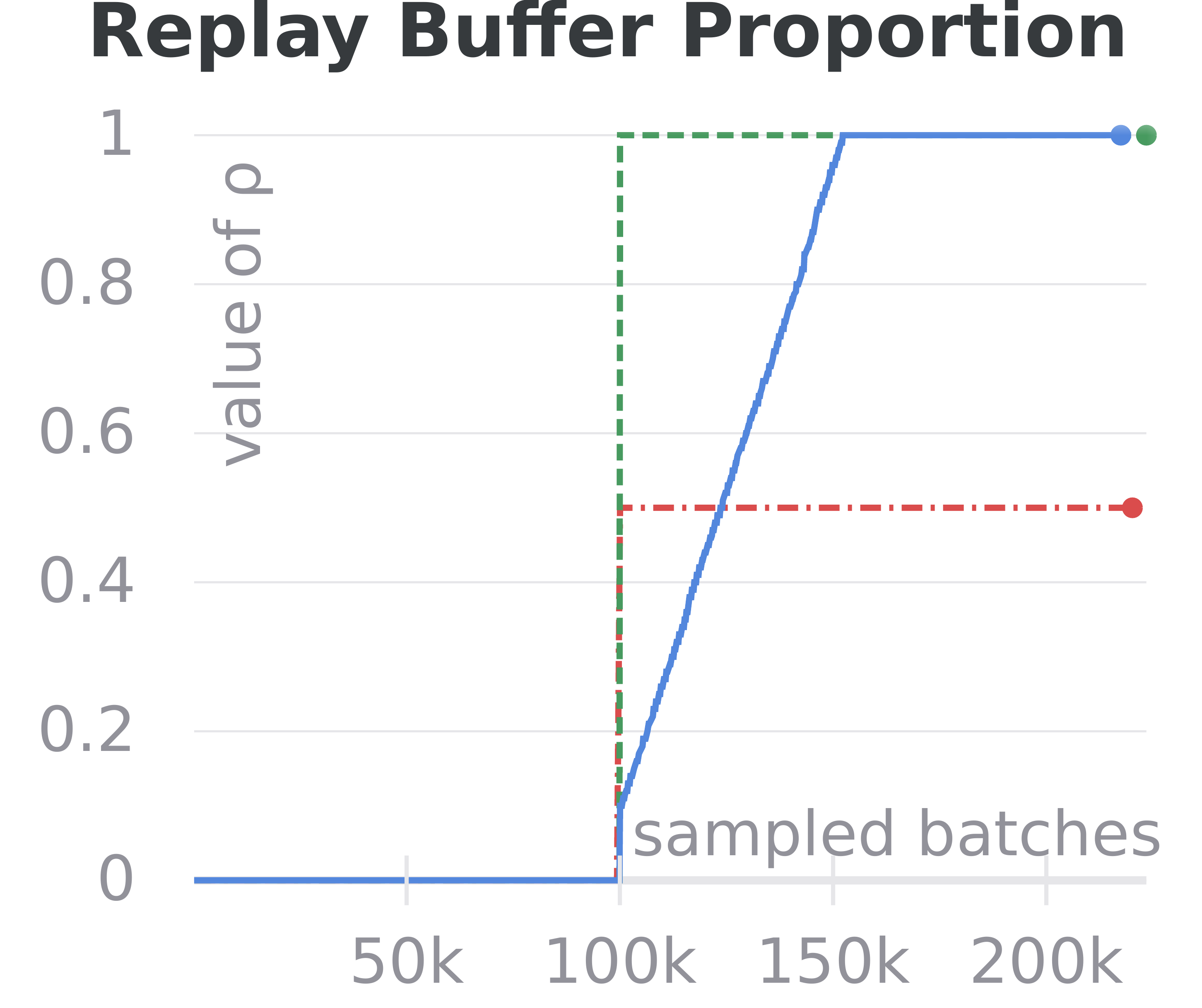}}
    \subfigure{\includegraphics[width=0.3\linewidth]{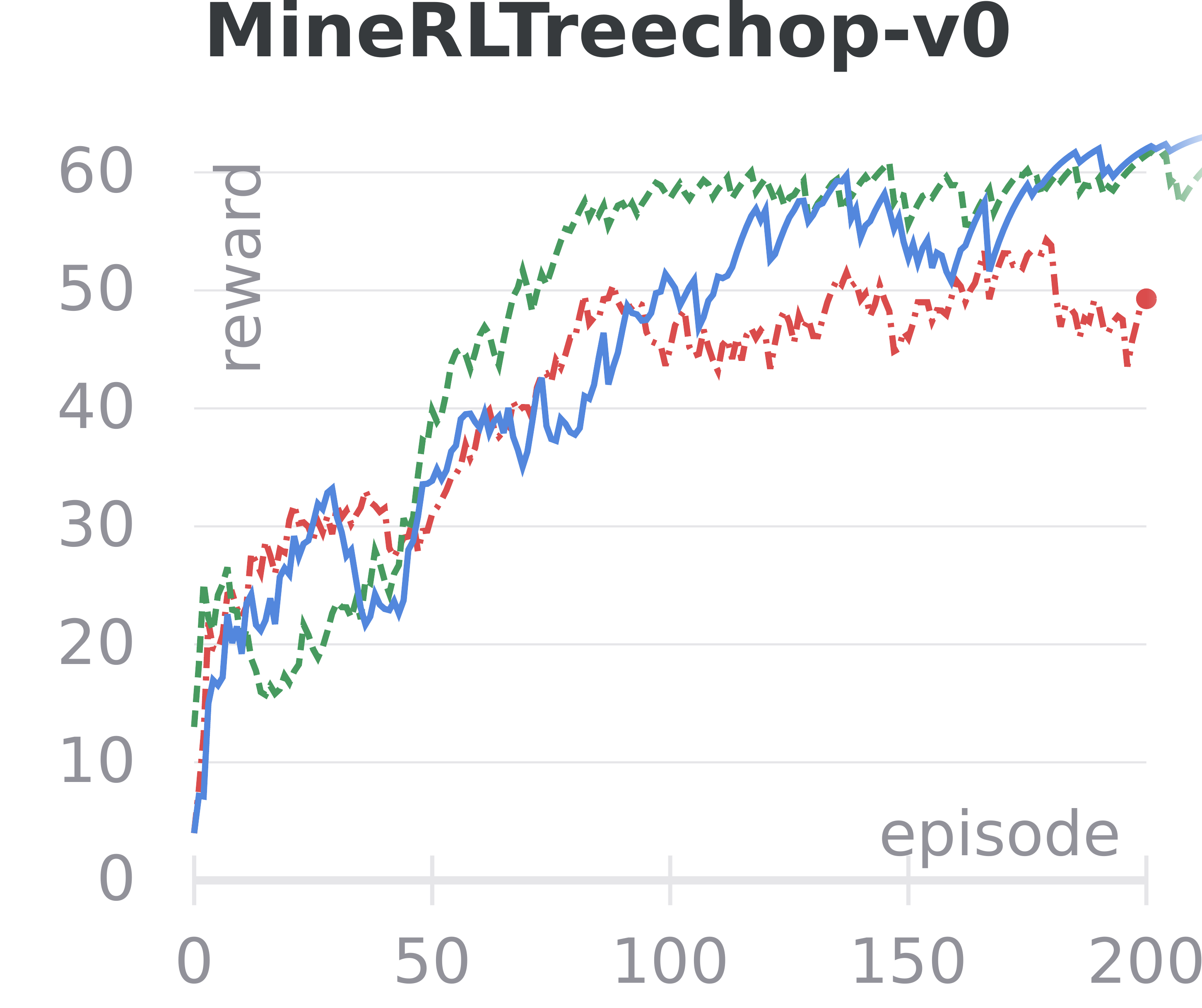}}
    \vspace{-0.2cm}
    \caption{Several variants of forgetting function $f_{rg}(k)$ realization. (\textbf{left}) TD loss curve for ForgER and DQfD agents. The first 100k sampled batches match the imitating phase. ForgER shows less TD loss then DQfD approach. (\textbf{middle}) The proportion $\rho$ of sampling expert and agent data in the replay buffer is indicated. (\textbf{right}) Total reward in \textit{Treechop} environment.} 
    \label{fig-overfitting}
\end{figure*}

\textbf{Discretization}. Discretization is a mapping between the continuous action space of the environment and discrete action space of the agent. But despite the fact that it lowers number of actions performed by the agent it also limits agents action space. Discretized actions from expert demonstrations may be inaccurate. In order to explore how forgetting affects the learning process with different discretization mappings we trained the agent for \textit{Treechop} with 7 and 10 actions and with different replay buffer structures. Each discretization was used with frameskip 4. The expert action is mapped to the agent’s action in the order shown in the tables 3 and 4 in Supplementary materials. The rotation angle is determined using the sum of 4 frames. For other actions, the most frequent was selected.

For each mapping we trained three configurations of the agent. The first configuration used forgetting expert data after 50 steps.
The second configuration used DQfD buffer.The third configuration had fixed proportion $\rho=0.1$ of the expert data in the batch.  Each version of the agent had 150000 pre-train steps and 250 episodes of training in the environment. Human demonstrations are considered as expert data.

As it can be seen from the graphs on Figures 11 when the discretization isn't accurate enough to map expert actions forgetting performs better, and when the discretization is good enough it doesn't perform worse than without forgetting.

\begin{figure*}[ht]
    \centering
    \subfigure{\includegraphics[trim={110px 20px 80px 50px},width=0.43\linewidth]{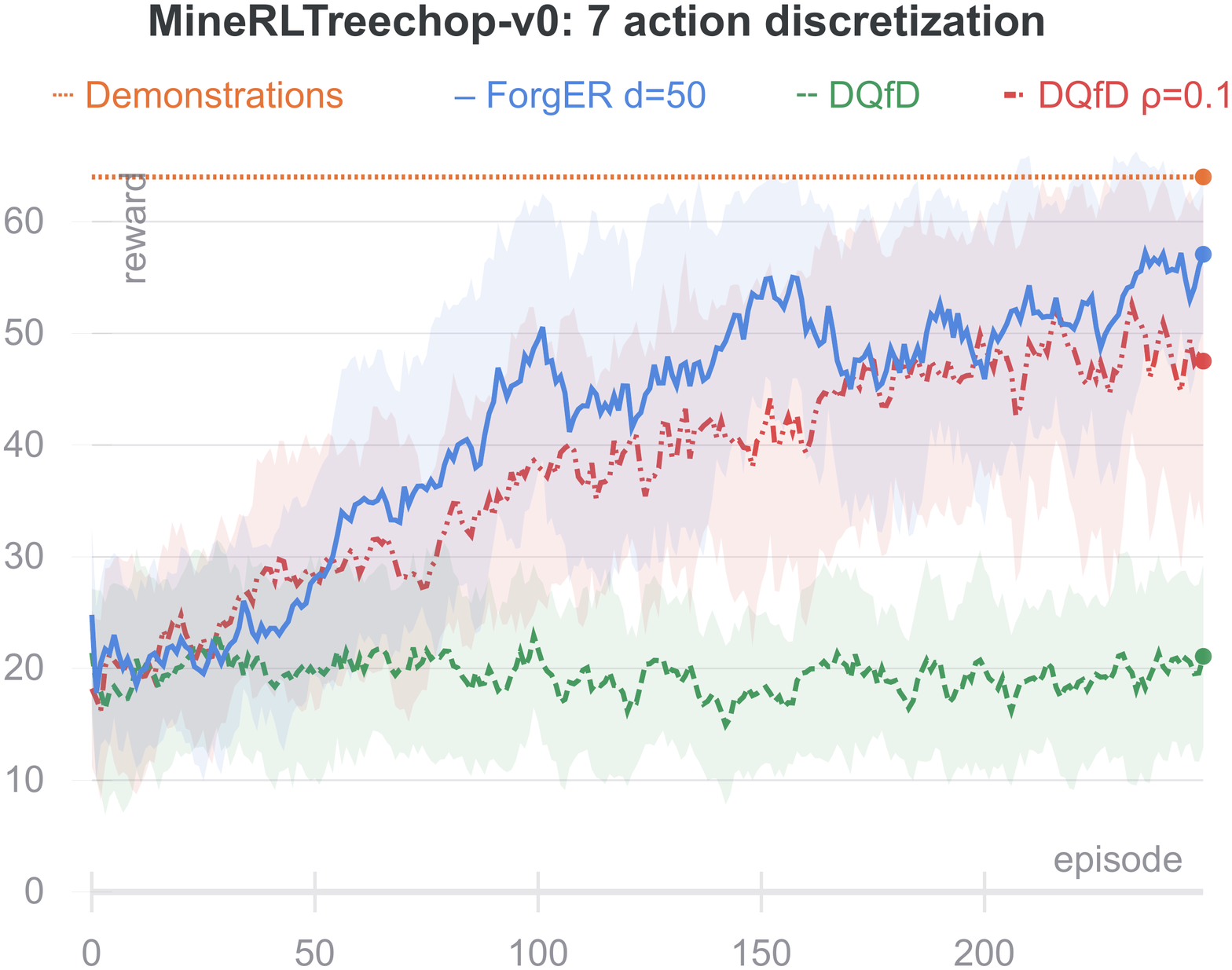}}
    \hspace{0.08\linewidth}
    \subfigure{\includegraphics[trim={110px 20px 80px 50px},width=0.43\linewidth]{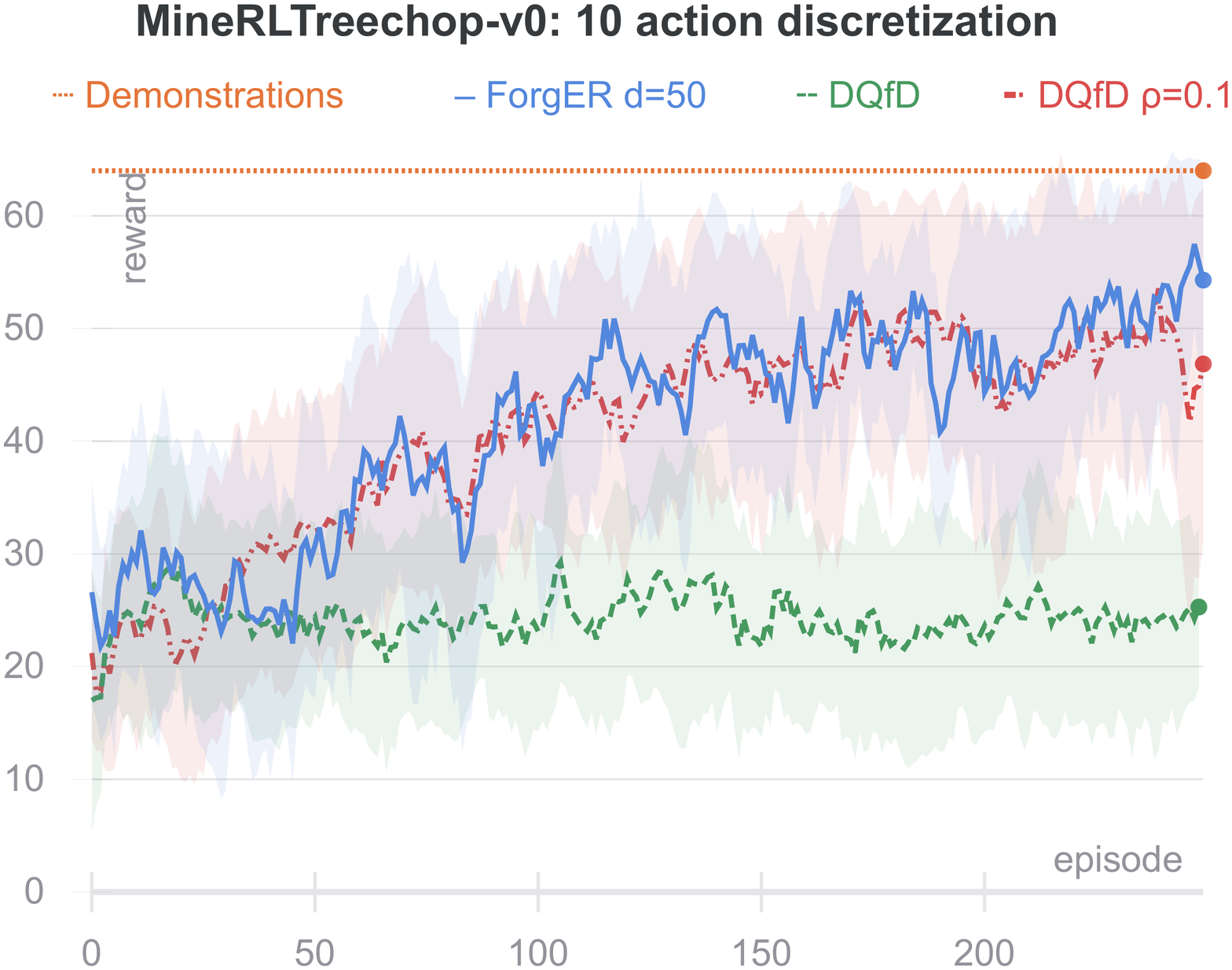}}
    \vspace{-0.2cm}
    \label{fig:exp:treechop}
    \caption{ \textbf{(left)} Mean episode reward for agents with discretetization with 7 actions. Forgetting is the agent that forgets expert data in 50 episodes. Fixed proportion is no forgetting configuration with fixed proportion of 0.1 of expert data used in batches.
    \textbf{(right)} Mean episode reward for agents with discretetization with 10 actions. Forgetting is the agent that forgets expert data in 50 episodes. Fixed proportion is no forgetting configuration with fixed proportion of $\rho=0.1$ of expert data used in batches. This is a better discretization, so that the difference between forgetting and fixed proportion agents is so insignificant.
    }
\end{figure*}

\textbf{Augmentations}. Task-specific augmentation was evaluated on \textit{Cobblestone} subtask in \textit{ObtainDiamond} (dense) environment and human demonstrations as expert data. It was compared with the version of the algorithm without augmentation. Both were averaged across 3 trials (see Figures 12). Agents for other subtasks had not been updated during evaluation. \textit{Cobblestone} agents had 50000 imitation steps and 50 episodes to forget expert demonstrations. Discretization with 10 actions for better behaviour cloning was used. Big dispersion can be explained with great variety of \textit{Minecraft} worlds, where the agent can appear. 

\begin{figure}[htp]
    \centering
    \includegraphics[width=8cm]{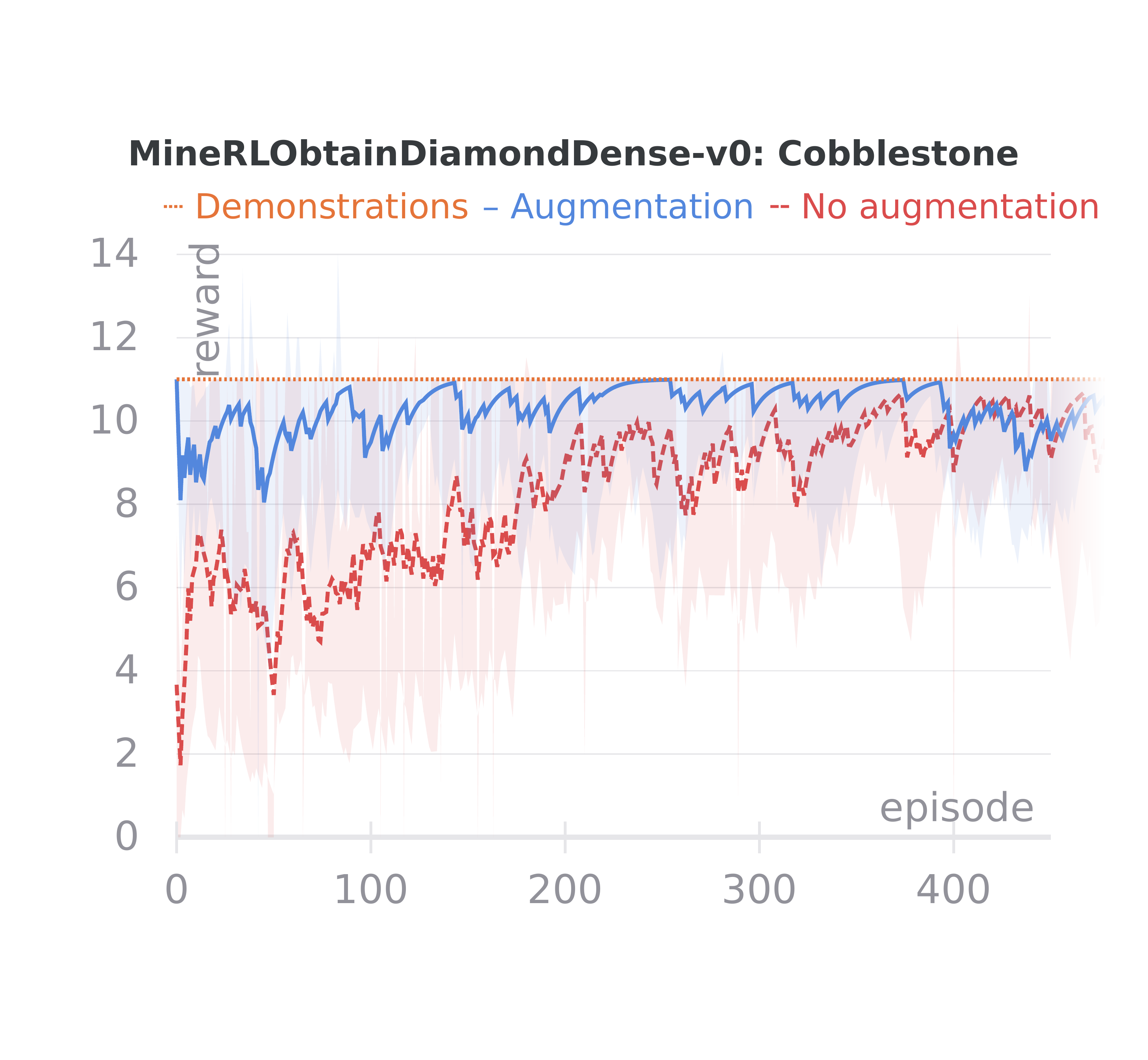}
    \caption{Mean episode reward for \textit{Cobblestone} subtask, in which agent must get 11 cobblestones. The reward for an episode can be more than 11 if the agent at the last moment picks up several cobblestones at once. Without augmentation it takes a lot of time for agent to understand where cobblestone is.}
    \label{fig:hie_aug}
\end{figure}

\textbf{Salience map indicated overfitting}. 
Paying attention to the crowns of trees when the agent is far from them is a more general strategy, as texture crowns always match, while trunk textures may vary. In most cases, the agent is focusing on the nearest tree. Imitation and DQfD agents pay attention to unimportant details: blocks that are not related to the tree chopping task, which indicates overfitting.

\begin{figure}[htp]
    \centering
    \includegraphics[width=1.0\linewidth]{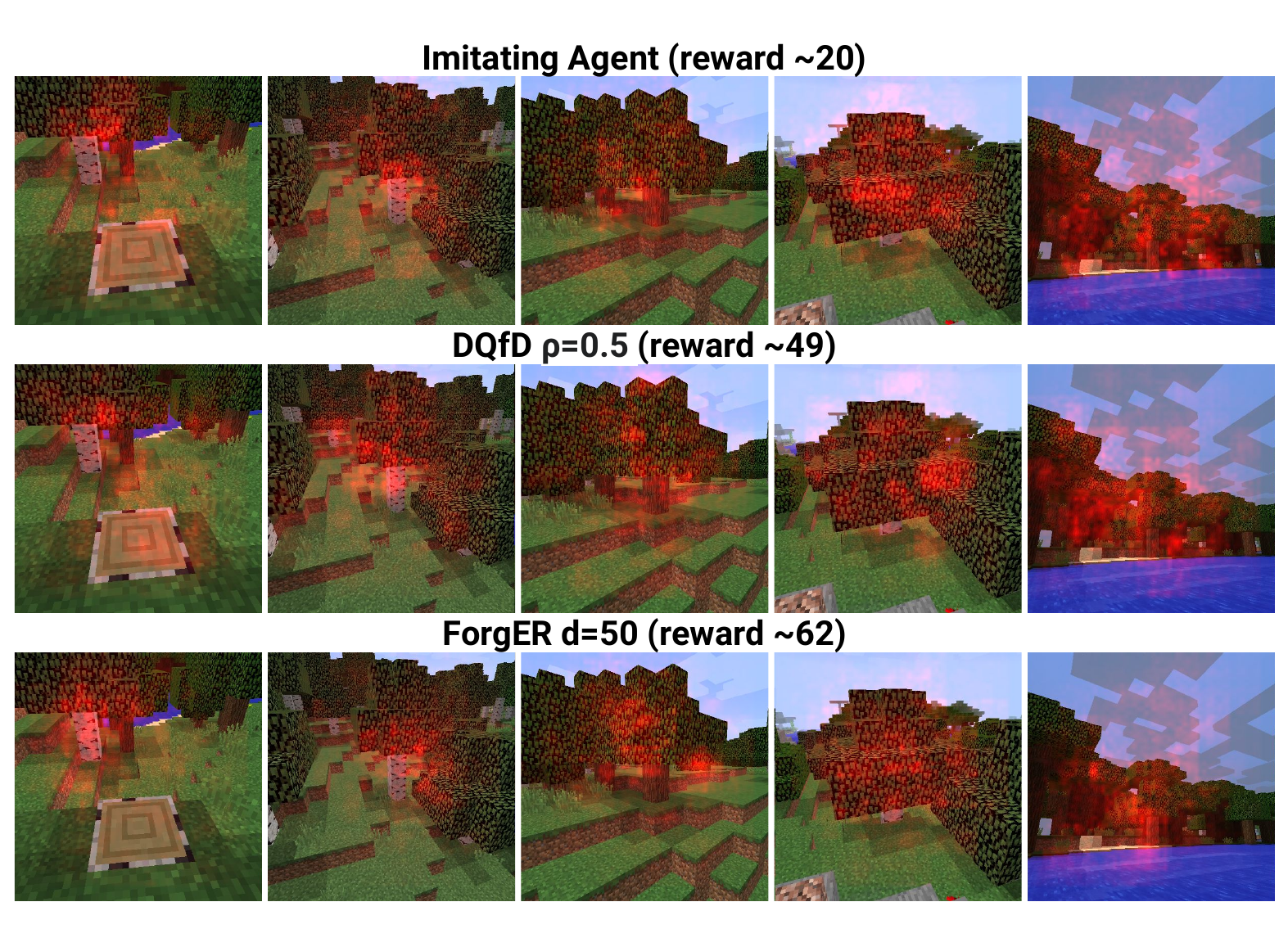}
    \caption{Salience analysis for imitating, DQfD and ForgER agents on \textit{Treechop} with the best discretization. The ForgER agent focusing on important details: crowns of trees when the agent is far from them and trunks when the agent is close. Noteworthy is the first frame on which ForgER, unlike other agents, does not pay attention to the stump. If the agent tries to cut it down, then it can dig itself down.  It is very difficult for an agent who fell underground to get to the surface, and there are no examples of such behavior in the demonstrations.}
    \label{fig:smap}
\end{figure}

\subsection{Hierarchical setting}
For the task of obtaining diamond in MineRL (\textit{hierarhical set}), we demonstrate the ForgER ability to use extracted subtask graph $G$ and use it for goal oriented ogranization of experience replay and data augmentation. In a crucial experiment, we compare our ForgER approach, ForgER++ modication (a heuristically modified hierarchy of subtasks $G$) with the best MineRL competition solution \cite{2019arXiv191208664S}. Because it was impossible to recreate the limitations of the competition; we reproduced the best solution without restrictions on the training time and number of steps.  Algorithms were tested on 1000 episodes using a common pool of 1000 seeds that were responsible for the procedural generation of the world and agent initial position. In addition to this experiment, we evaluated ForgER++ approach on 1500 episodes. This approach allowed the agent to mine diamond for the first time in the MineRL competition released. As a result, our approach surpassed the best solution of the MineRL competition, which, in its turn, was the best among all solutions based on DQfD, PPO, GAIL and other algorithms.

\subsection{Subtask graph extraction in MineRL}

For the task of obtaining diamond in MineRL, we propose the following automatic approach for subtasks extraction. We consider the time of item appearance in the inventory in chronological order. Sequential items of the same type are combined into an item with quantity. The final subgoal tree can be a sequence obtained from a single trajectory. An example of a marked sequence of subtasks presented here: \textit{log}(6), \textit{planks}(24), \textit{crafting table}(1), \textit{sticks}(4), \textit{wooden pickaxe}(1), \ldots, \textit{iron pickaxe}(1).

For the obtaining diamond task, the tree diminishes to a chain of subtasks. Each subtask is to obtain the required amount of indicated resource. The agent receives a pseudo-reward when it obtains an item related to the current subtask (for example $+1$, for receiving $1$ \textit{log}). In addition to pseudo reward in the environment, pseudo rewards are added to expert data. We also used the fact that actions to craft items are strongly related to subtasks. Craft actions aimed at solving the subtask were included in each subtask.

Table~\ref{MineRL} shows the results of testing three algorithms: the best algorithm presented at the competition (trained from scratch), FrogER with an automatically extracted chain of subtasks and ForgER with a modified chain of subtasks (ForgER++).

\begin{table}[htp]
    \caption{The table shows the results of testing the algorithms on 1000 evaluation episodes. For items is shown the number of episodes in which the agent obtained a reward for receiving it. The first column shows the results of the best solution in the MineRL competition.}
    \label{MineRL}
    \centering
    \begin{tabular}{llll}
        \toprule

        item & MineRL & ForgER & ForgER++  \\
        \midrule
        log& 859& \textbf{882}& 867\\
        planks& 805& \textbf{806}& 792\\
        stick& 718& 747& \textbf{790}\\
        crafting table& 716& 744& \textbf{790}\\
        wooden pickaxe& 713& 744& \textbf{789}\\
        cobblestone& 687& 730& \textbf{779}\\
        stone pickaxe& 642& 698& \textbf{751}\\
        furnace& 19& 48& \textbf{98}\\
        iron ore& 96& 109& \textbf{231}\\
        iron ingot& 19& 48& \textbf{98}\\
        iron pickaxe& 12& 43& \textbf{83}\\
        diamond& 0& 0& \textbf{1}\\
        \midrule
        mean reward& 57.701& 74.09& \textbf{104.315}\\
        \bottomrule
    \end{tabular}
\end{table}

We used several task-specific settings in ForgER and ForgER++. We add a small white noise (mean = 0, std = 0.6) to the camera rotation actions, which improves both exploration and behavior of policies after imitating phase. Also we use the \textit{log} subtask policy learnt on the auxiliary \textit{TreeChop} environment.

\subsection{Hyperparameters}

In the table~\ref{table:params:common} we show the parameters for ForgER approach used in all experiments. The table~\ref{table:params:both} represents parameters, which differ for visual and vector environments. The value for the number of pre-training steps varies for each of the environments.

\begin{table}[!htb]
    \caption{Shared parameters used for both environment sets.}
    \label{table:params:common}
    \centering
    \begin{tabular}{ll}
        \toprule
        parameter & value  \\
        \midrule
        N-step return weight $\lambda_1$ & 1.0 \\
        Margin loss weight $\lambda_2$ & 1.0 \\
        L2 regularization weight $\lambda_3$ & $10^{-5}$ \\
        Expert margin $l$ & 0.4 \\
        $\epsilon$-greedy initial & 0.1 \\ 
        $\epsilon$-greedy final & 0.01 \\ 
        $\epsilon$-greedy decay & 0.99 \\ 
        Prioritized replay exponent $\alpha$ & 0.4 \\
        Prioritized replay constant $\epsilon_a$ & 0.0001 \\
        Prioritized replay constant $\epsilon_d$ & 1.0 \\
        Prioritized replay importance sampling exponent $\beta_0$ & 0.6 \\
        n-step return & 10 \\
        batch size & 32 \\
        \bottomrule \\
    \end{tabular}
\end{table}

\begin{table}[!htb]
    \caption{Parameters used for \textit{simple} and \textit{visual} environments.}
    \label{table:params:both}
    \centering
    \begin{tabular}{lll}
        \toprule
        parameter & value for \textit{simple set} & value for \textit{visual set}  \\
        \midrule
        Target network update period $\tau$ & 2,000 & 10,000 \\ 
        Use noisy layers & False & True \\
        Agent replay buffer capacity & 100,000 & 450,000 \\ 
        \bottomrule \\
    \end{tabular}
\end{table}
\section{Related Work}
Some components in ForgER are not new. Sampling strategy from buffer potentially could deal a lot of learning process problems. In prioritized experience replay~\cite{PER} authors achieve that by picking samples from buffer regarding how often these samples are used for backpropagation in the network based on the values of their temporal difference (TD) loss function.

The idea to store the human experience in additional expert buffer to increase performance in difficult Atari games was proposed in Human Experience Replay~\cite{DBLP:journals/corr/HosuR16}. More than 5 hours of game play was stored and was sampled as a 16 out of 32 examples in the training batch without any pre-train and supervised loss. Due to modified expirience replay, authors for the first time achieve better than random agent performance in Montezuma's Revenge.

A combination of TD and classification losses in a batch algorithm in a model-free setting are the key components in RL with Expert Demonstrations (RLED)~\cite{RLED}. DQfD differs from RLED in that agent is pre-trained on the demonstration data initially and the batch of self-generated data grows over time and is issued as experience replay to train deep Q-networks. In addition, a prioritized replay mechanism is used to balance the amount of demonstration data in each mini-batch~\cite{DBLP:journals/corr/HesterVPLSPSDOA17}. As for buffer, DQfD uses only one, both for expert and agent trajectories. All of these techniques provide the next level of performance in tasks like Montezuma's Revenge and stays as a state of the art algorithm in learning from demonstration. We use the same approach for learning but with different strategy of experience replay buffer structuring that helped achieve a significant increase in efficiency. 

There are also implementations of algorithms that learn from demonstration based not on DQN, but on policy optimization methods. One of them is Normalized Actor Critic (NAC)~\cite{gao2018reinforcement}, which is almost entirely based on Soft Actor Critic~\cite{haarnoja2018soft}. However it has a special normalizing gradient supplement, which allows not to overfit on demonstrations, due to the fact that expert data is often imperfect. As it turns out, NAC doesn't work well in vision-based environments. Policy Optimization with Demonstrations (POfD)~\cite{kang2018policy} is an on-policy algorithm, the main difference from Generative Adversarial Imitation Learning (GAIL)~\cite{DBLP:journals/corr/HoE16} is that GAIL optimizes the policy to confuse the discriminator, whereas POfD has a demonstration-guided exploration term in learning objection. POfD showed massive improvement over GAIL, using a low number of expert data, and proved that it is not biased by the imperfect data. Despite all of that, GAIL based algorithms are very computationally heavy and they are very sensitive to changes in the action space.

Most recent research in the field of learning from demonstration is a Recurrent Replay Distributed DQN from Demonstrations (R2D3)~\cite{R2D3}, that outperformed multiple state of the art baselines. The authors extend the R2D2 algorithm and add expert data buffer. In conclusion, it was mentioned, that the ratio of expert data and agent data is one of the key parameters of their algorithm and finetuning of it could significantly increase results. R2D3 used only a fixed ratio and established that a small demo ratio is the most desirable case. We have extend this investigation and suggest more appropriate ways to treat demo-ratio as a decreasing variable and were able to adapt the approach of using demonstrations to the hierarchical case.
\section{Conclusion}
We presented ForgER, a novel algorithm for reinforcement learning from demonstrations in complex partially observable environments including hierarchical settings. We propose task-oriented structure of the experience replay buffer with embedded procedure of forgetting imperfect expert trajectories. By exploiting the hierarchical structure of the demonstrations in case of its availability, we can obtain hierarchical policies that generalize substantially better than SOTA methods. Additionally, we structure the replay buffer by using augmented data on the imitating phase for each task specified policy. With ForgER, we have achieved two main goals: we attained real high sample efficiency by combining a hierarchical approach and using demonstration data, and at the same time we reduced the quality requirements for this expert trajectories.

We experimentally investigate the ForgER techniques and showed the features of its implementation in the case of different sources of imperfectness in expert data. In our experiments, we demonstrated that ForgER can solve very complex hierarchical vision-based environments such as \textit{Minecraft} where we solve the main problem of diamond obtaining in MineRL setting. In the MineRL competition, various tricks were used to adapt the known approaches to the hierarchical POMDP environment, and only our complex original method helped to achieve the main goal of this competition and surpass all other solutions.

Broadly, we have shown that ForgER outperforms existing RL approaches based on expert demonstrations, especially those with high-dimensional inputs and hierarchical complex goals. While we used a simple sequential graph of subgoals, in future work we aim to explore how more complex subgoal structures with loops can be automatically detected and how it can improve imitation phase. In addition, we believe that the main idea of using forgetting experience replay will open doors to incorporating more sophisticated memory-based techniques into RL.

\section*{Broader Impact}
We believe that our ForgER method of reinforcement learning using demonstrations that may be non-expert and noisy, significantly reduces the computational threshold for entering this area of research. Our method opens the way for small teams that do not have access to huge computing resources, as in large corporations, to participate and create RL algorithms that also show impressive results, for example, in environments such as Minecraft. We consider this the main positive broader outcome. Negative outcome can be attributed to lowering the threshold for getting any results from unmarked data, which can potentially be used for some bad purposes. However, this can be attributed to almost any new technology in the field of AI.

\bibliographystyle{abbrvnat}
\bibliography{preprint}

\begin{thebibliography}{34}
\providecommand{\natexlab}[1]{#1}
\providecommand{\url}[1]{\texttt{#1}}
\expandafter\ifx\csname urlstyle\endcsname\relax
  \providecommand{\doi}[1]{doi: #1}\else
  \providecommand{\doi}{doi: \begingroup \urlstyle{rm}\Url}\fi

\bibitem[Andrychowicz et~al.(2017)Andrychowicz, Wolski, Ray, Schneider, Fong,
  Welinder, McGrew, Tobin, Abbeel, and Zaremba]{andrychowicz2017hindsight}
M.~Andrychowicz, F.~Wolski, A.~Ray, J.~Schneider, R.~Fong, P.~Welinder,
  B.~McGrew, J.~Tobin, O.~P. Abbeel, and W.~Zaremba.
\newblock Hindsight experience replay.
\newblock In \emph{Advances in neural information processing systems}, pages
  5048--5058, 2017.

\bibitem[Brockman et~al.(2016)Brockman, Cheung, Pettersson, Schneider,
  Schulman, Tang, and Zaremba]{brockman2016openai}
G.~Brockman, V.~Cheung, L.~Pettersson, J.~Schneider, J.~Schulman, J.~Tang, and
  W.~Zaremba.
\newblock Openai gym.
\newblock \emph{arXiv preprint arXiv:1606.01540}, 2016.

\bibitem[Cassandra et~al.(1994)Cassandra, Kaelbling, and
  Littman]{cassandra1994acting}
A.~R. Cassandra, L.~P. Kaelbling, and M.~L. Littman.
\newblock Acting optimally in partially observable stochastic domains.
\newblock In \emph{AAAI}, volume~94, pages 1023--1028, 1994.

\bibitem[Duan et~al.(2017)Duan, Andrychowicz, Stadie, Ho, Schneider, Sutskever,
  Abbeel, and Zaremba]{duan2017one}
Y.~Duan, M.~Andrychowicz, B.~Stadie, O.~J. Ho, J.~Schneider, I.~Sutskever,
  P.~Abbeel, and W.~Zaremba.
\newblock One-shot imitation learning.
\newblock In \emph{Advances in neural information processing systems}, pages
  1087--1098, 2017.

\bibitem[Eysenbach et~al.(2019)Eysenbach, Salakhutdinov, and
  Levine]{eysenbach2019search}
B.~Eysenbach, R.~R. Salakhutdinov, and S.~Levine.
\newblock Search on the replay buffer: Bridging planning and reinforcement
  learning.
\newblock In \emph{Advances in Neural Information Processing Systems}, pages
  15220--15231, 2019.

\bibitem[Gao et~al.(2018)Gao, Lin, Yu, Levine, Darrell,
  et~al.]{gao2018reinforcement}
Y.~Gao, J.~Lin, F.~Yu, S.~Levine, T.~Darrell, et~al.
\newblock Reinforcement learning from imperfect demonstrations.
\newblock \emph{arXiv preprint arXiv:1802.05313}, 2018.

\bibitem[Guss et~al.(2019)Guss, Codel, Hofmann, Houghton, Kuno, Milani,
  Mohanty, Liebana, Salakhutdinov, Topin, et~al.]{minerlcomp}
W.~H. Guss, C.~Codel, K.~Hofmann, B.~Houghton, N.~Kuno, S.~Milani, S.~Mohanty,
  D.~P. Liebana, R.~Salakhutdinov, N.~Topin, et~al.
\newblock The {M}ine{RL} competition on sample efficient reinforcement learning
  using human priors.
\newblock \emph{NeurIPS Competition Track}, 2019.

\bibitem[H.~Van~Hasselt and Silver(2016)]{DoDQN}
A.~G. H.~Van~Hasselt and D.~Silver.
\newblock Deep reinforcement learning with double q-learning.
\newblock In \emph{Proceedings of the 30th AAAI Conference on Artificial
  Intelligence (AAAI)}, pages 2094--2100, 2016.

\bibitem[Haarnoja et~al.(2018)Haarnoja, Zhou, Abbeel, and
  Levine]{haarnoja2018soft}
T.~Haarnoja, A.~Zhou, P.~Abbeel, and S.~Levine.
\newblock Soft actor-critic: Off-policy maximum entropy deep reinforcement
  learning with a stochastic actor.
\newblock \emph{CoRR}, abs/1801.01290, 2018.
\newblock URL \url{http://arxiv.org/abs/1801.01290}.

\bibitem[Hester et~al.(2017{\natexlab{a}})Hester, Vecer{\'i}k, Pietquin,
  Lanctot, Schaul, Piot, Horgan, Quan, Sendonaris, Osband, Dulac-Arnold,
  Agapiou, Leibo, and Gruslys]{DQfD}
T.~Hester, M.~Vecer{\'i}k, O.~Pietquin, M.~Lanctot, T.~Schaul, B.~Piot,
  D.~Horgan, J.~Quan, A.~Sendonaris, I.~Osband, G.~Dulac-Arnold, J.~Agapiou,
  J.~Z. Leibo, and A.~Gruslys.
\newblock Deep q-learning from demonstrations.
\newblock In \emph{AAAI}, 2017{\natexlab{a}}.

\bibitem[Hester et~al.(2017{\natexlab{b}})Hester, Vecer{\'{\i}}k, Pietquin,
  Lanctot, Schaul, Piot, Sendonaris, Dulac{-}Arnold, Osband, Agapiou, Leibo,
  and Gruslys]{DBLP:journals/corr/HesterVPLSPSDOA17}
T.~Hester, M.~Vecer{\'{\i}}k, O.~Pietquin, M.~Lanctot, T.~Schaul, B.~Piot,
  A.~Sendonaris, G.~Dulac{-}Arnold, I.~Osband, J.~Agapiou, J.~Z. Leibo, and
  A.~Gruslys.
\newblock Learning from demonstrations for real world reinforcement learning.
\newblock \emph{CoRR}, abs/1704.03732, 2017{\natexlab{b}}.
\newblock URL \url{http://arxiv.org/abs/1704.03732}.

\bibitem[Ho and Ermon(2016)]{DBLP:journals/corr/HoE16}
J.~Ho and S.~Ermon.
\newblock Generative adversarial imitation learning.
\newblock \emph{CoRR}, abs/1606.03476, 2016.
\newblock URL \url{http://arxiv.org/abs/1606.03476}.

\bibitem[Hosu and Rebedea(2016)]{DBLP:journals/corr/HosuR16}
I.~Hosu and T.~Rebedea.
\newblock Playing atari games with deep reinforcement learning and human
  checkpoint replay.
\newblock \emph{CoRR}, abs/1607.05077, 2016.
\newblock URL \url{http://arxiv.org/abs/1607.05077}.

\bibitem[Johnson et~al.(2016)Johnson, Hofmann, Hutton, and
  Bignell]{johnson2016malmo}
M.~Johnson, K.~Hofmann, T.~Hutton, and D.~Bignell.
\newblock The malmo platform for artificial intelligence experimentation.
\newblock In \emph{IJCAI}, pages 4246--4247, 2016.

\bibitem[{Kalashnikov} et~al.(2018){Kalashnikov}, {Irpan}, {Pastor}, {Ibarz},
  {Herzog}, {Jang}, {Quillen}, {Holly}, {Kalakrishnan}, {Vanhoucke}, and
  {Levine}]{2018arXiv180610293K}
D.~{Kalashnikov}, A.~{Irpan}, P.~{Pastor}, J.~{Ibarz}, A.~{Herzog}, E.~{Jang},
  D.~{Quillen}, E.~{Holly}, M.~{Kalakrishnan}, V.~{Vanhoucke}, and S.~{Levine}.
\newblock {QT-Opt: Scalable Deep Reinforcement Learning for Vision-Based
  Robotic Manipulation}.
\newblock \emph{arXiv e-prints}, art. arXiv:1806.10293, Jun 2018.

\bibitem[Kang et~al.(2018)Kang, Jie, and Feng]{kang2018policy}
B.~Kang, Z.~Jie, and J.~Feng.
\newblock Policy optimization with demonstrations.
\newblock In \emph{International Conference on Machine Learning}, pages
  2469--2478, 2018.

\bibitem[Mart{\'\i}nez et~al.(2017)Mart{\'\i}nez, Alenya, and
  Torras]{martinez2017relational}
D.~Mart{\'\i}nez, G.~Alenya, and C.~Torras.
\newblock Relational reinforcement learning with guided demonstrations.
\newblock \emph{Artificial Intelligence}, 247:\penalty0 295--312, 2017.

\bibitem[Mnih and Kavukcuoglu(2016)]{A3C}
A.~P. M. M. G. A. L. T. H. T. S.~D. Mnih, V.;~Badia and K.~Kavukcuoglu.
\newblock Asynchronous methods for deep reinforcement learning.
\newblock In \emph{In International Conference on Machine Learning (ICML)},
  pages 1928--1937, 2016.

\bibitem[Paine et~al.(2019)Paine, Gulcehre, Shahriari, Denil, Hoffman, Soyer,
  Tanburn, Kapturowski, Rabinowitz, Williams, et~al.]{R2D3}
T.~L. Paine, C.~Gulcehre, B.~Shahriari, M.~Denil, M.~Hoffman, H.~Soyer,
  R.~Tanburn, S.~Kapturowski, N.~Rabinowitz, D.~Williams, et~al.
\newblock Making efficient use of demonstrations to solve hard exploration
  problems.
\newblock \emph{arXiv preprint arXiv:1909.01387}, 2019.

\bibitem[Piot et~al.(2014)Piot, Geist, and Pietquin]{RLED}
B.~Piot, M.~Geist, and O.~Pietquin.
\newblock Boosted bellman residual minimization handling expert demonstrations.
\newblock In \emph{Machine Learning and Knowledge Discovery in Databases},
  pages 549--564, Berlin, Heidelberg, 2014. Springer Berlin Heidelberg.

\bibitem[Piot and Pietquin(2014)]{Piot2014}
M.~Piot, B.;~Geist and O.~Pietquin.
\newblock Boosted bellman residual minimization handling expert demonstrations.
\newblock In \emph{EuropeanConference on Machine Learning (ECML)}, 2014.

\bibitem[Puterman(2014)]{MDP}
M.~L. Puterman.
\newblock \emph{Markov Decision Processes: Discrete Stochastic Dynamic
  Programming}.
\newblock John Wiley \& Sons, 2014.

\bibitem[Rajeswaran et~al.(2017)Rajeswaran, Kumar, Gupta, Vezzani, Schulman,
  Todorov, and Levine]{rajeswaran2017learning}
A.~Rajeswaran, V.~Kumar, A.~Gupta, G.~Vezzani, J.~Schulman, E.~Todorov, and
  S.~Levine.
\newblock Learning complex dexterous manipulation with deep reinforcement
  learning and demonstrations.
\newblock \emph{arXiv preprint arXiv:1709.10087}, 2017.

\bibitem[Schaul and Silver(2016)]{PER}
J.~A.~I. Schaul, T.;~Quan and D.~Silver.
\newblock Prioritized experience replay.
\newblock In \emph{Proceedings ofthe International Conference on Learning
  Represen- tations}, 2016.

\bibitem[Shu et~al.(2017)Shu, Xiong, and Socher]{shu2017hierarchical}
T.~Shu, C.~Xiong, and R.~Socher.
\newblock Hierarchical and interpretable skill acquisition in multi-task
  reinforcement learning.
\newblock \emph{arXiv preprint arXiv:1712.07294}, 2017.

\bibitem[{Skrynnik} et~al.(2019){Skrynnik}, {Staroverov}, {Aitygulov},
  {Aksenov}, {Davydov}, and {Panov}]{2019arXiv191208664S}
A.~{Skrynnik}, A.~{Staroverov}, E.~{Aitygulov}, K.~{Aksenov}, V.~{Davydov}, and
  A.~I. {Panov}.
\newblock {Hierarchical Deep Q-Network from Imperfect Demonstrations in
  Minecraft}.
\newblock \emph{arXiv e-prints}, art. arXiv:1912.08664, Dec 2019.

\bibitem[Sutton and Barto(2018)]{Sutton}
R.~S. Sutton and A.~G. Barto.
\newblock \emph{Reinforcement learning: An introduction}.
\newblock MIT press, 2018.

\bibitem[Sutton et~al.(1999)Sutton, Precup, and Singh]{sutton1999between}
R.~S. Sutton, D.~Precup, and S.~Singh.
\newblock Between mdps and semi-mdps: A framework for temporal abstraction in
  reinforcement learning.
\newblock \emph{Artificial intelligence}, 112\penalty0 (1-2):\penalty0
  181--211, 1999.

\bibitem[V.~Mnih and et~al.(2015)]{DQN}
D.~S. V.~Mnih, K.~Kavukcuoglu and et~al.
\newblock Human-level control through deep reinforcement learning.
\newblock \emph{Nature}, 518\penalty0 (7540):\penalty0 529--533, 2015.

\bibitem[{Vecerik} et~al.(2017){Vecerik}, {Hester}, {Scholz}, {Wang},
  {Pietquin}, {Piot}, {Heess}, {Roth{\"o}rl}, {Lampe}, and
  {Riedmiller}]{2017arXiv170708817V}
M.~{Vecerik}, T.~{Hester}, J.~{Scholz}, F.~{Wang}, O.~{Pietquin}, B.~{Piot},
  N.~{Heess}, T.~{Roth{\"o}rl}, T.~{Lampe}, and M.~{Riedmiller}.
\newblock {Leveraging Demonstrations for Deep Reinforcement Learning on
  Robotics Problems with Sparse Rewards}.
\newblock \emph{arXiv e-prints}, art. arXiv:1707.08817, Jul 2017.

\bibitem[Watkins and Dayan(1992)]{Qlearn}
C.~J. C.~H. Watkins and P.~Dayan.
\newblock Q-learning.
\newblock \emph{Machine learning}, 8\penalty0 (3-4):\penalty0 279--292, 1992.

\bibitem[{Yarats} et~al.(2019){Yarats}, {Zhang}, {Kostrikov}, {Amos}, {Pineau},
  and {Fergus}]{2019arXiv191001741Y}
D.~{Yarats}, A.~{Zhang}, I.~{Kostrikov}, B.~o. {Amos}, J.~{Pineau}, and
  R.~{Fergus}.
\newblock {Improving Sample Efficiency in Model-Free Reinforcement Learning
  from Images}.
\newblock \emph{arXiv e-prints}, art. arXiv:1910.01741, Oct 2019.

\bibitem[{Zhu} et~al.(2018){Zhu}, {Wang}, {Merel}, {Rusu}, {Erez}, {Cabi},
  {Tunyasuvunakool}, {Kram{\'a}r}, {Hadsell}, {de Freitas}, and
  {Heess}]{2018arXiv180209564Z}
Y.~{Zhu}, Z.~{Wang}, J.~{Merel}, A.~{Rusu}, T.~{Erez}, S.~{Cabi},
  S.~{Tunyasuvunakool}, J.~{Kram{\'a}r}, R.~{Hadsell}, N.~{de Freitas}, and
  N.~{Heess}.
\newblock {Reinforcement and Imitation Learning for Diverse Visuomotor Skills}.
\newblock \emph{arXiv e-prints}, art. arXiv:1802.09564, Feb 2018.

\bibitem[Ziyu~Wang and de~Freitas(2016)]{DuDQN}
M.~H. H. v. H. M.~L. Ziyu~Wang, Tom~Schaul and N.~de~Freitas.
\newblock Dueling network architectures for deep reinforcement learning.
\newblock In \emph{Proceedings of the 33nd International Conference on Machine
  Learning (ICML 2016)}, pages 1995--2003, 2016.

\end{thebibliography}

\end{document}